%% file: main.tex
\documentclass{article}

\usepackage{microtype}
\usepackage{graphicx}
\usepackage{subfigure}
\usepackage{booktabs} 
\input{tex/package}
\usepackage{hyperref}

\usepackage[accepted]{mlsys2025}

\mlsystitlerunning{RepV: Scalable Neurosymbolic Verification}

\begin{document}

\twocolumn[
\mlsystitle{RepV: Safety-Separable Latent Spaces for Scalable Neurosymbolic Plan Verification}



\mlsyssetsymbol{equal}{*}

\begin{mlsysauthorlist}
\mlsysauthor{Yunhao Yang}{to}
\mlsysauthor{Neel P.~Bhatt}{to}
\mlsysauthor{Pranay Samineni}{to}
\mlsysauthor{Rohan Siva}{to}
\mlsysauthor{Zhangyang Wang}{to}
\mlsysauthor{Ufuk Topcu}{to}
\end{mlsysauthorlist}

\mlsysaffiliation{to}{The University of Texas at Austin, United States}

\mlsyscorrespondingauthor{Yunhao Yang}{yunhaoyang234@utexas.edu}
\mlsyscorrespondingauthor{Neel P. Bhatt}{npbhatt@utexas.edu}

\mlsyskeywords{Machine Learning, MLSys}

\vskip 0.3in

\input{tex/00-abstract}

]



\printAffiliationsAndNotice{} 

\input{tex/01-introduction}
\input{tex/02-related_work}
\input{tex/03-preliminary}
\input{tex/04-method}
\input{tex/05-demonstration}
\input{tex/05-quantitative}

\input{tex/06-conclusion}

\bibliography{reference}
\bibliographystyle{mlsys2025}

\input{tex/A-appendix}

\end{document}

%% file: tex/package.tex
\usepackage{subcaption}
\usepackage{newfloat}
\usepackage{adjustbox}
\usepackage{listings}
\usepackage[table]{xcolor}
\usepackage{tikz}
\usepackage{mathtools} 
\usepackage{amsfonts} %
\usepackage{svg}
\usepackage{bbm}
\usepackage{pgfplots}
\pgfplotsset{compat=1.18}
\usetikzlibrary[pgfplots.groupplots]

\usepackage{wrapfig}

\usetikzlibrary{arrows, automata, positioning}
\tikzset{auto, >=stealth}
\tikzset{every edge/.append style={shorten >= 1pt}}
\tikzset{
    main node/.style={circle,draw,minimum size=1cm,inner sep=0pt},
}

\DeclareCaptionStyle{ruled}{labelfont=normalfont,labelsep=colon,strut=off} 
\usepackage{listings}
\definecolor{codegreen}{rgb}{0,0.6,0}
\definecolor{codegray}{rgb}{0.5,0.5,0.5}
\definecolor{codepurple}{rgb}{0.58,0,0.82}
\definecolor{backcolour}{rgb}{0.95,0.95,0.92}
\definecolor{figurepurple}{rgb}{0.631,0.333,0.906}

\lstdefinestyle{code}{
    backgroundcolor=\color{backcolour},   
    commentstyle=\color{codegreen},
    keywordstyle=\color{magenta},
    numberstyle=\tiny\color{codegray},
    stringstyle=\color{codepurple},
    basicstyle=\ttfamily\footnotesize,
    breakatwhitespace=false,         
    breaklines=true,                 
    captionpos=b,                    
    keepspaces=true,                 
    numbers=left,                    
    numbersep=5pt,                  
    showspaces=false,                
    showstringspaces=false,
    showtabs=false,                  
    tabsize=2,
    moredelim=**[is][\color{blue}]{@}{@}
}

\lstdefinestyle{base}{
  language=tcl,
  emptylines=1,
  breaklines=true,
  basicstyle=\ttfamily\color{blue},
  moredelim=**[is][\color{purple}]{@}{@},
}

\usepackage{graphicx, array, blindtext}
\usepackage{dblfloatfix}

\usepackage{colortbl}

\definecolor{codebg}{RGB}{245,245,245}
\definecolor{codeframe}{RGB}{200,200,200}
\definecolor{codecomment}{RGB}{0,128,0}
\definecolor{codekeyword}{RGB}{0,0,180}
\definecolor{codestring}{RGB}{163,21,21}

\lstdefinestyle{nicePython}{
    language=Python,
    basicstyle=\ttfamily\footnotesize,
    backgroundcolor=\color{codebg},
    frame=single,
    rulecolor=\color{codeframe},
    frameround=tttt,
    showstringspaces=false,
    commentstyle=\color{codecomment}\itshape,
    keywordstyle=\color{codekeyword}\bfseries,
    stringstyle=\color{codestring},
    tabsize=4,
    belowskip=1em,
    aboveskip=1em,
    breaklines=true,
    breakatwhitespace=true,
    columns=fullflexible,
}

%% file: tex/00-abstract.tex
\begin{abstract}
As AI systems migrate to safety–critical domains, verifying that their actions comply with well-defined rules remains a challenge.  
Formal methods provide provable guarantees but demand hand-crafted temporal-logic specifications, offering limited expressiveness and accessibility.
Deep learning approaches enable evaluation of plans against natural-language constraints, yet their opaque decision process invites misclassifications with potentially severe consequences. 
We introduce \texttt{RepV}, a neurosymbolic verifier that unifies both views by \emph{learning a latent space where safe and unsafe plans are linearly separable}.
Starting from a modest seed set of plans labeled by an off-the-shelf model checker, \texttt{RepV} trains a lightweight projector that embeds each plan, together with a language model-generated rationale, into a low-dimensional space; a frozen linear boundary then verifies compliance for \emph{unseen} natural-language rules in a single forward pass.

Beyond binary classification, \texttt{RepV} provides a \emph{probabilistic guarantee} on the likelihood of correct verification based on its position in the latent space. This guarantee enables a \emph{guarantee-driven refinement} of the planner, improving rule compliance without human annotations.
Empirical evaluations show that \texttt{RepV} improves compliance prediction accuracy by up to 15\% compared to baseline methods while adding fewer than 0.2 M parameters. Furthermore, our refinement framework outperforms ordinary fine-tuning baselines across various planning domains.
These results show that safety-separable latent spaces offer a scalable, plug-and-play primitive for reliable neurosymbolic plan verification. Code and data are available at: \href{https://repv-project.github.io/}{\textcolor{blue}{https://repv-project.github.io/}}
\end{abstract}

%% file: tex/01-introduction.tex
\section{Introduction}

As AI systems take on greater roles in safety-critical applications, ensuring their actions comply with well-defined rules remains challenging. Compliance is a foundation for safety, reliability, and accountability across healthcare \cite{osifowokan2025regulatory, sbodio2024collaborative, daram2025explainable}, finance \cite{deshpande2024regulatory, balakrishnan2024leveraging}, and autonomous systems \cite{yazdanpanah2023reasoning, he2021challenges}. Traditional rule-based systems enable formal verification against specified logic-based constraints \cite{grosan2011rule}. In contrast, AI systems, driven by deep learning models, lack interpretability and formal guarantees due to their black-box nature, which increases the difficulty of rule compliance verification.

Formal methods and language models provide two approaches for verifying compliance, yet each has its limitations. Existing works have extended formal methods techniques to support compliance verification across multiple input modalities, including natural language and programming languages \cite{Yang2022AutomatonBasedRO, li2024formal, yang2024joint}. 
Although these methods offer guarantees, the logical specifications they rely on have limited expressiveness and demand domain expertise, limiting their accessibility. 
On the other hand, several works query language models for compliance classification, allowing inference with natural language rules \cite{guan2024deliberative, hassani2024enhancing, ying2024automatic}. Still, they can suffer from misclassifications due to the black-box nature of language models, hence limiting their reliability.

\begin{figure*}[t]
    \centering
    \includegraphics[width=0.75\linewidth]{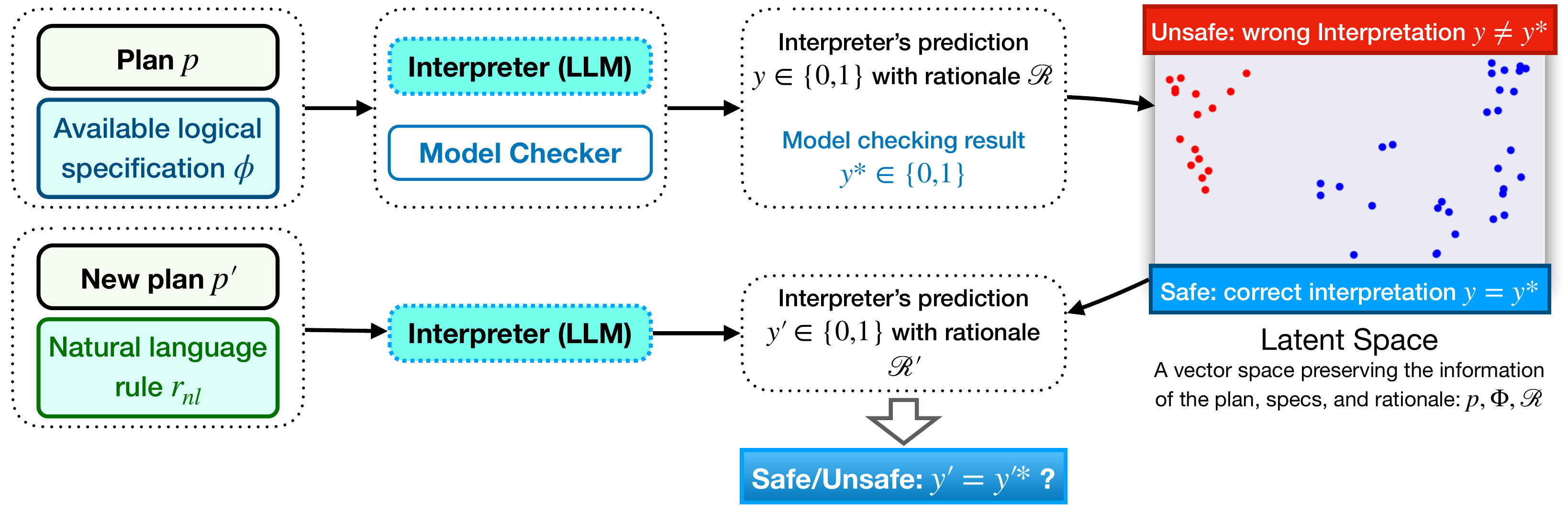}
    \vspace{-10pt}
    \caption{A high-level illustration of \texttt{RepV}. It first learns a latent space where the safe (the interpreter's prediction aligns with the model checker) and unsafe plan representations are separable. Then, it classifies the new plan's compliance against the new natural language rules based on the spatial location of the plan representation in the latent space.}
    \label{fig: logic}
    \vspace{-10pt}
\end{figure*}

This work addresses the challenge of predicting rule compliance in planning tasks, where interpretibility and reliability are essential, but often in tension. We consider AI systems that generate and execute a plan, i.e., a sequence of actions, subject to a set of constraints. These constraints may be combinations of logical specifications and natural language rules. While logical specifications support formal verification through tools such as model checkers \cite{Pnueli77LTL, baier2008principles}, they require structured input and domain expertise. Conversely, natural language rules are easy to obtain and express but lack the semantics required for formal verification.

We introduce \texttt{RepV}, a neurosymbolic verifier that bridges the gap between the two means of expressing constraints and verifies whether a plan complies with constraints, during system execution. 
\texttt{RepV} begins with a modest set of plans labeled via formal verification tools. It then trains a lightweight projector that maps each plan, along with an LLM-generated rationale, into a low-dimensional latent space. 
Within this space, each point represents the \emph{alignment} between the interpreter’s predicted compliance and the formal verification outcome. We define a plan as \emph{safe} when this alignment holds, i.e. when the interpreter’s prediction matches the verification result.
Once the projector is trained, \texttt{RepV} classifies compliance with unseen natural language rules based on a plan’s spatial location in the latent space.
We present an illustration of \texttt{RepV} in Figure \ref{fig: logic}.

Building on the alignment-based latent space, we establish a \emph{probabilistic guarantee} (statistically calibrated probability estimate) on \texttt{RepV}'s correctness of verification. Instead of producing a binary label, the verifier quantifies the likelihood that its inferred compliance outcome aligns with the formal verification outcome.
This likelihood is calibrated using the distance between the generated plan and the cluster centroid of a correctly aligned representation in latent space.
Rather than providing a binary assessment of the plan, \texttt{RepV} offers a calibrated confidence estimate, enabling uncertainty-aware compliance assessments.

We further leverage the probabilistic guarantee to guide refinement of the external \emph{planner (multimodal foundation model)} in a data-efficient manner. We propose two approaches for \emph{guarantee-driven refinement}: Supervised fine-tuning and preference optimization.
The former identifies high-guarantee compliant plans and uses them to fine-tune the planner's parameters, consolidating reliable planning behaviors.
The latter ranks data preferences via the probabilistic guarantee and train the planner to produce the preferred planning behaviors.
Together, the guarantee-driven refinement forms a lightweight feedback loop that continuously improves the plan-generation model without additional human labeling or full retraining.

Empirical evaluations show that \texttt{RepV} improves compliance prediction accuracy by up to 15\% compared to baseline methods while adding fewer than 0.2 M parameters. The evaluations are over four domains: simulated driving, outdoor navigation (i.e., real-world driving), indoor navigation, and 3D aerial navigation. \texttt{RepV} achieves over 90\% prediction accuracy in all domains, showcasing its reliability in safety-critical domains and generalizability to unseen environments and rules.

Furthermore, the guarantee-driven refinement improves the planner’s ability to generate rule-compliant plans and outperforms ordinary fine-tuning baselines.
Across four robotic domains, this refinement process improves the planner’s plan-generation compliance rate by up to 15\% while reducing convergence time by more than half compared to ordinary supervised fine-tuning.
The results demonstrate that probabilistic guarantees provide an efficient mechanism for continual planner improvement, transforming verification feedback into actionable learning signals.

\textbf{Contributions:} We develop a neurosymbolic verifier, \texttt{RepV}, that enables accessible and reliable planning through three key contributions: \textbf{\textit{(1)}} \texttt{RepV} enables compliance verification with natural language rules, eliminating the need to handcraft logical specifications. \textbf{\textit{(2)}} It establishes a probabilistic guarantee that quantifies the likelihood of each compliance judgment, enabling uncertainty-aware verification with natural-language rules. \textbf{\textit{(3)}} The probabilistic guarantee enables continual improvement of external plan generation models with minimal overhead.

%% file: tex/02-related_work.tex
\section{Related Work}

\textbf{Formal Methods for Compliance Verification: }
Formal methods offer rigorous guarantees for verifying system behavior against well-defined specifications \cite{mehdipour2023formal}. Techniques such as model checking have been widely applied in safety-critical systems \cite{baier2008principles}. Recent works have expanded these techniques to support more input modalities \cite{bhatt2024know}, such as programming languages and natural language, by translating them into structured (mathematical) representations \cite{li2024formal, yang2024joint, yang2024reasoning}. While compelling in well-defined domains, these techniques struggle with the ambiguity and variability of rules expressed in natural language.

\textbf{Language Models for Compliance Checking: }
Recent advancements in language models with strong reasoning capabilities enable new approaches to compliance checking \cite{guan2024deliberative, ying2024automatic, gan2024large}. These models extract obligations and detect non-compliance in legal, financial, and healthcare texts, outperforming formal methods approaches in flexibility and scalability \cite{hafizi2024auditing, hassani2024enhancing, hassani2024rethinking, berger2023towards}. While these approaches effectively enhance model safety, lacking formal guarantees can be unreliable, especially in safety-critical domains.

\textbf{Neurosymbolic Approaches for Compliance Checking: }
Several studies integrate learned symbolic representations of behaviors from logic-based systems to assess whether the behaviors comply with well-defined rules \cite{barbara2023neuro, pakina2024neuro, paul2025neuro, ahn2025towards, bhatt2025vln, daggitt2024vehicle}. Despite promising results, they require handcrafted logic-based systems or are limited to structured domains. Hence, they lack scalability towards long-term development. Additionally, several works use LLM-based modules to transform informal constraints into formal logic \cite{ganguly2024proofthoughtneurosymbolic, Lee_2025}, improving scalability. However, the reliability of this transformation is not guaranteed due to the black box LLM.

\textbf{Latent Signals from LLM Activations: } Recent studies have revealed that LLMs encode rich internal signals in their activations (intermediate outputs), such as indicators of errors~\cite{orgad2024llms, karnik2025preemptivedetectionsteeringllm, bhatt2025uncap}, hallucinations~\cite{ferrando2024know}, refusals~\cite{arditi2024refusal}, and overthinking tendencies~\cite{chen2025seal}. They often capture subtle model behaviors that are not directly observable from the output text. Inspired by these findings, we explore whether safety-related signals can be detected within the model's encoded activations.

%% file: tex/03-preliminary.tex
\section{Preliminaries}
\label{sec: prelim}

\textbf{Formal Methods in Sequential Decision-Making: }
Sequential decision-making refers to the process by which an agent selects actions over time, where each action affects future states. In this work, we focus on \emph{planning problems} which are a subset of sequential decision-making problems.
Given a planning task, we define a \emph{plan} as a sequence of actions that achieve a task goal. In this work, we describe a plan in either natural language or programming language. 

Formal methods provide tools for modeling and verifying planning problems represented mathematically. Suppose we have a \emph{transition system} $TS$ modeling the world knowledge or autonomous systems, a \emph{finite-state automaton} (FSA) $\mathcal{A}$ representing the plan, and a \emph{temporal logic specification} $\phi$ constraining the temporal ordering and logical relations between actions.
$\mathcal{A}$ and $TS$ are mathematical structures with a finite number of states \cite{baier2008principles}, and $\phi$ is a logical formula with temporal operators such as ``always'' and ``eventually'' \cite{Pnueli77LTL}.

We apply a \emph{model checker}, a tool from formal methods, to verify whether $\mathcal{A}$, when implemented in $TS$, satisfies $\phi$, denoted as $\mathcal{A} \otimes TS \models \phi$ \ \cite{baier2008principles, Cimatti2002NuSMV}. We present the formal definitions of these terminologies in Appendix \ref{app: definition}.

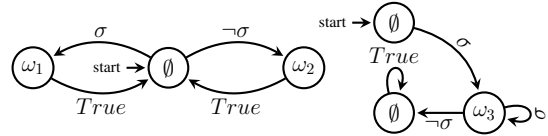
\begin{figure}[t]
    \centering
    \input{figure/automata/if-else}
    \caption{Automaton corresponding to a syntax \textit{``if $\sigma$, do $\omega_1$, else $\omega_2$''} (left) and \textit{``while $\sigma$, do $\omega_3$''} (right). }
    \vspace{-15pt}
    \label{fig: if-else-automaton}
\end{figure}

\textbf{Text-Based Plan to Automaton: }
To formally verify a plan against logic specifications, we need to express the plan in a formal representation such as an FSA. Existing works have developed methods for converting natural language or programming language to FSA \cite{Yang2022AutomatonBasedRO, yang2024joint}.

We use an algorithm, denoted as \texttt{L2A}, that takes a text-based plan $p$, expressed in natural language or programming language, as input and converts it into an FSA $\mathcal{A} = \texttt{L2A}(p)$. In particular, the algorithm parses the text input into phrases (e.g., keywords and variables in a programming language). It then follows pre-defined grammar to convert the parsed phrases into automaton states and transitions.
We present an automaton conversion example in Figure \ref{fig: if-else-automaton} and present the details in Appendix \ref{app: text2aut}.

%% file: figure/automata/if-else.tex
\begin{tikzpicture}[thick,scale=.6, node distance=2.2cm, every node/.style={transform shape}]
	\node[initial, state] (0) at (0, 0) {\Large $\emptyset$};
	\node[state] (1) at (-3, 0) {\Large $\omega_1$};
	\node[state] (2) at (3, 0) {\Large $\omega_2$};

        \draw[->, shorten >=1pt] (0) to[bend left] node[above, align=center, sloped] {\Large $\neg \sigma$} (2);
        \draw[->, shorten >=1pt] (2) to[ bend left] node[below, align=center, sloped] {\Large $True$} (0);
        \draw[->, shorten >=1pt] (0) to[bend right] node[above, align=center, sloped] {\Large $\sigma$} (1);
        \draw[->, shorten >=1pt] (1) to[ bend right] node[below, align=center, sloped] {\Large $True$} (0);

    \node[initial, state] (3) at (5, 1) {\Large $\emptyset$};
	\node[state] (4) at (7, -1) {\Large $\omega_3$};
    \node[state] (5) at (5, -1) {\Large $\emptyset$};

        \draw[->, shorten >=1pt] (3) to[bend left] node[above, align=center, sloped] {\Large $\sigma$} (4);
        \draw[->, shorten >=1pt] (4) to[loop right] node[below, align=center, sloped] {\Large $\sigma$} ();
        \draw[->, shorten >=1pt] (4) to[] node[below, align=center, sloped] {\Large $\neg \sigma$} (5);
        \draw[->, shorten >=1pt] (5) to[loop above] node[above, align=center, sloped] {\Large $True$} ();
\end{tikzpicture}

%% file: tex/04-method.tex
\begin{figure*}[t]
    \centering
    \includegraphics[width=0.85\linewidth]{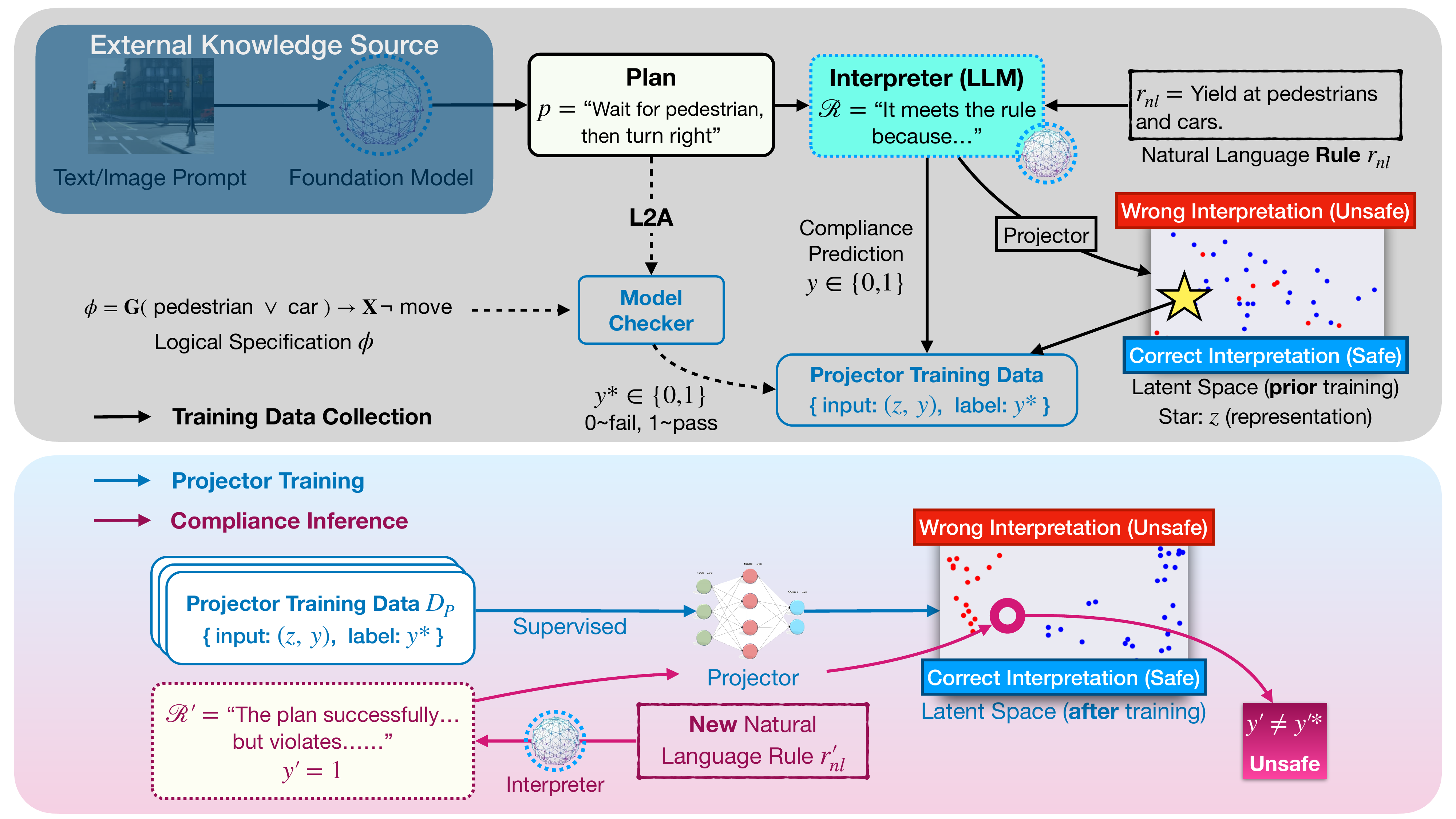}
    \vspace{-10pt}
    \caption{An illustration for collection of auto-labeled training data for the projector, training the projector to construct a safety-separable latent space, and predicting plan compliance against natural language rules via the spatial location of the projector's output representation.}
    \label{fig: pipeline}
    \vspace{-10pt}
\end{figure*}

\section{RepV: A Neurosymbolic Verifier}
\label{sec: method}

\texttt{RepV} is a neurosymbolic verifier that unifies formal verification with representation learning to \textbf{enable verification of externally generated action plans against natural language rules}. At its core, \texttt{RepV} constructs a latent space where safe and unsafe plans are linearly separable, enabling efficient inference of compliance with natural-language rules. This latent-space embedding bridges the symbolic rigor with linguistic accessibility, establishing a foundation to extend verification toward probabilistic guarantees and iterative refinement.

\textbf{Components: }
\texttt{RepV} comprises four main components: an external knowledge source for plan generation, an interpreter $\mathcal{M_I}$, a projector $\mathcal{P}$, and a model checker $\mathcal{M_C}$.

\textit{- Knowledge Source (Planner): } An external source used for plan generation. In this setting, we use a \emph{multimodal foundation model} that accepts text- or image-based task prompts as input, generating a text-based plan that fulfills the task. Note that the \textit{rule} is a part of the task prompt as well. For brevity, we denote it as a \emph{planner}.

\textit{- Interpreter: } A language model $\mathcal{M_I}$ that processes a generated plan $p$ and a natural language rule $r_{nl}$, producing a binary classification $y \in \{0,1\}$ indicating its prediction on rule compliance (1 means comply) and a text-based rationale $\mathcal{R}$. Mathematically, we have $y, \mathcal{R} = \mathcal{M_I}(p, r_{nl})$. An illustrative example of interpreter outputs is in Section \ref{sec: exp-setup}.

\textit{- Projector: } A multi-layer perception $\mathcal{P}: \mathbb{R}^{\epsilon} \rightarrow \mathbb{R}^m$ that takes the $\epsilon$-dimensional \emph{text embeddings} $emb(p, \mathcal{R}) \in \mathbb{R}^{\epsilon}$ derived from the plan and the interpreter’s rationale and maps them into an $m$-dimensional latent space, where compliance and noncompliance can be distinguished spatially.

\textit{- Model Checker: } A formal verification tool $\mathcal{M_C}$ verifies whether a plan $p$, represented as a finite-state automaton $\mathcal{A}_p = \texttt{L2A}(p)$, satisfies a logical specification $\phi$ when executed in a world/system model $TS$.
It returns a binary ground-truth label indicating compliance: 
\begin{center}
    $\quad \mathcal{M_C}(\mathcal{A}_p, \phi) = 1 \text{ if } \mathcal{A}_p \otimes TS \models \phi, \text{ else }0$.
\end{center}

\textbf{Definition of Safety: }
To learn a safety-separable latent space, we first define the terminology on \emph{safe} and \emph{unsafe}.

\textbf{Definition 1:}
Given a plan $p$, a natural language rule $r_{nl}$, and a logical specification $\phi$ that corresponds to $r_{nl}$ (i.e., $r_{nl}$ describes the semantic meaning of $\phi$), we obtain a natural language interpretation $y, \mathcal{R} = \mathcal{M_I}(p, r_{nl})$ and the formal verification outcome $y^{*} = \mathcal{M_C}(\mathcal{A}_p, \phi)$.

We define the plan $p$ as \textbf{safe} with respect to the rule $r_{nl}$ if and only if its predicted classification aligns with the formal verification outcome:
\begin{equation}
p \text{ is safe } \iff y = y^{*}.
\end{equation}

Note that the “safety” in \texttt{RepV} differs from the everyday notion of physical or behavioral safety. A plan is considered safe even if it violates the rule ($y^{*} = 0$), as long as the interpreter’s prediction correctly reflects that violation ($y = 0$).
Conversely, a plan is unsafe when there is a misalignment between the interpreter’s prediction and the verified outcome ($y\neq y^{*}$), regardless of whether the plan itself satisfies or violates the rule.
Hence, safety in this context measures \emph{semantic consistency between reasoning modalities}—the alignment of linguistic interpretation and formal verification.

\textbf{Problem Statement: } Given a plan $p$ extracted from external sources (e.g., AI-generated) and a natural language rule $r_{nl}$, \textbf{verify} whether each plan complies with the rule, denoted as $p \models r_{nl}$, and estimate the \textbf{probability} of such verification outcome being correct.

\subsection{Learning a Safety-Separable Latent Space}

To achieve reliable verification across linguistic and formal modalities, \texttt{RepV} learns a latent space where semantically \emph{safe} and \emph{unsafe} plans become linearly separable. This latent space serves as a bridge between linguistic reasoning (from $\mathcal{M_I}$) and symbolic verification (from $\mathcal{M_C}$), enabling the system to infer compliance for new natural-language rules based on spatial proximity.

\textbf{Training Data Collection:}
We derive each training sample from a triplet $(p, r_{nl}, \phi)$, where $p$ is a generated plan, $r_{nl}$ is a natural-language rule, and $\phi$ is its corresponding logical specification. The interpreter $\mathcal{M_I}$ predicts a compliance label $y$ and generates a rationale $\mathcal{R}$:
\begin{equation}
y, \mathcal{R} = \mathcal{M_I}(p, r_{nl}).
\end{equation}
We then convert the plan $p$ into an automaton $\mathcal{A}_p = \texttt{L2A}(p)$ and verify it against $\phi$ by the model checker $\mathcal{M_C}$, producing a ground-truth label $y^{*}$:
\begin{equation}
y^{*} = \mathcal{M_C}(\mathcal{A}_p, \phi).
\end{equation}

\textbf{Latent Representation:}
We obtain a text embedding $emb(p, \mathcal{R}) \in \mathbb{R}^{\epsilon}$ using a pretrained text encoder $emb()$ that encodes the plan $p$ and the interpreter’s rationale $\mathcal{R}$. 

A lightweight projector $\mathcal{P}: \mathbb{R}^{\epsilon} \rightarrow \mathbb{R}^{m}$ maps this embedding into a low-dimensional latent representation $z = \mathcal{P}(emb(p, \mathcal{R}))$. Each representation $z$ is associated with a binary label $y_{safe}$ that indicates whether the interpreter’s prediction aligns with the verified outcome:
\begin{equation}
y_{safe} = 
\begin{cases}
1, & y = y^{*} \quad \text{(safe)},\\
0, & y \neq y^{*} \quad \text{(unsafe)}.
\end{cases}
\end{equation}

\textbf{Projector Optimization:}
We then train the projector using a cross-entropy loss that minimizes the classification error between the predicted safety $\hat{y}_{safe}$ (predicted via the spatial location in the latent space) and the verified label $y_{safe}$:
\begin{equation}
\mathcal{L}_{proj} = - \sum_{i} y_{safe}^{(i)} \log \hat{y}_{safe}^{(i)} + (1 - y_{safe}^{(i)}) \log (1 - \hat{y}_{safe}^{(i)}).\nonumber
\end{equation}
Through supervised learning, \texttt{RepV} learns a safety-separable latent space $\mathcal{Z} \subset \mathbb{R}^m$, where spatial clusters encode alignment between linguistic interpretation and formal verification. As shown in Figure~\ref{fig: pipeline}, the process involves collecting formally verified samples, projecting them into $\mathcal{Z}$, and learning a linear boundary that partitions safe and unsafe representations.

\begin{figure*}[t]
    \centering
    \includegraphics[width=0.85\linewidth]{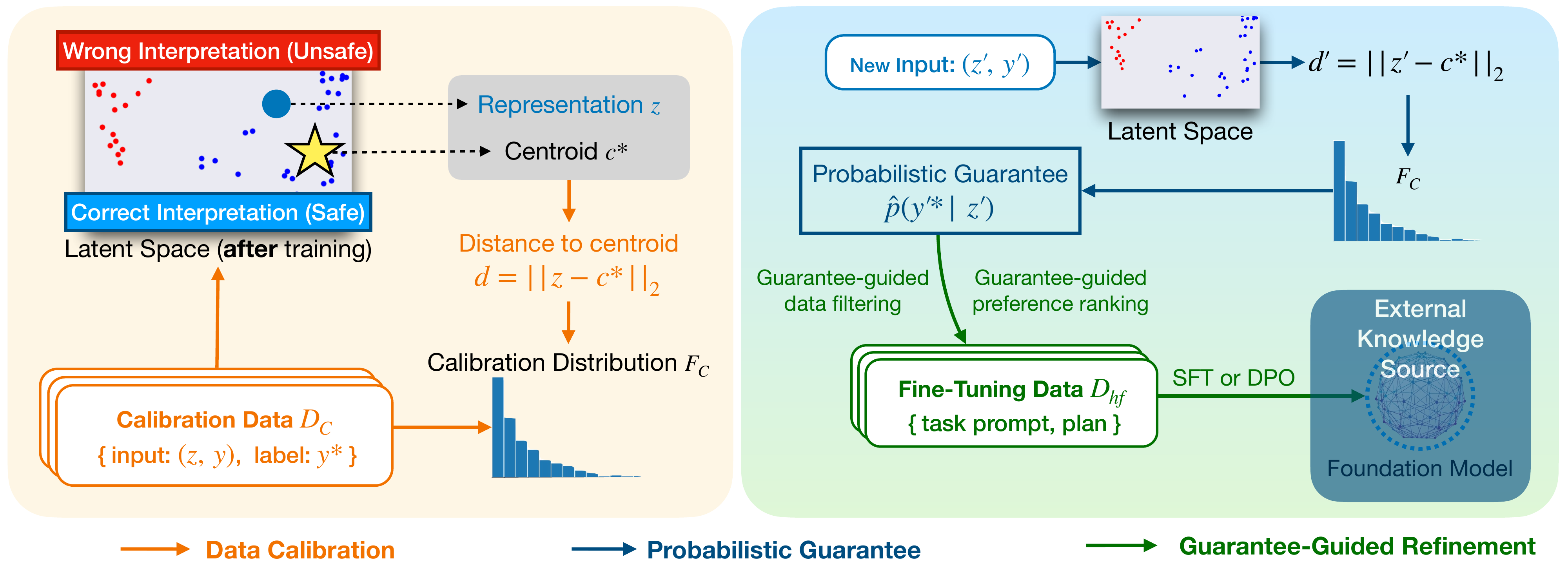}
    \caption{Overview of the probabilistic verification and refinement pipeline. \textbf{Left:} After training the latent space, we extract a calibration distribution $F_C$ using distances from correctly verified samples to their cluster centroids. \textbf{Right:} Given a new plan–rule pair, we compute the distance between its latent representation and the nearest centroid $c^*$. Then, we use $F_C$ to compute the probabilistic guarantee. The guarantee can be subsequently used as a feedback signal for upgrading the planner.}
    \label{fig: pipeline-refine}
\end{figure*}

\subsection{Probabilistic Rule Verification}

After obtaining the latent space $\mathcal{Z}$, \texttt{RepV} infers compliance for new plans and rules by projecting them into $\mathcal{Z}$ and estimating a probabilistic guarantee that reflects the probability of correct verification.

\textbf{Inference Procedure:}
Given a new plan $p'$ and a natural language rule $r'_{nl}$, the interpreter $\mathcal{M_I}$ produces a binary classification and rationale:
\begin{equation}
y', \mathcal{R}' = \mathcal{M_I}(p', r'_{nl}), \nonumber
\end{equation}
and the projector maps the text embedding into the latent space:
\begin{equation}
z' = \mathcal{P}(emb(p', \mathcal{R}')). \nonumber
\end{equation}
We then attach a linear classifier $C: \mathbb{R}^m \rightarrow \{0,1\}$ to predict the compliance label:
\begin{equation}
\hat{y}'_{safe} = C(z') \in \{0,1\}, \nonumber
\end{equation}
where $1$ indicates that the interpreter’s reasoning is aligned with the expected formal outcome.

\textbf{Calibration Dataset:}
To provide a probability estimate for this inference, we maintain a small calibration set $\mathcal{D}_c = \{(z_i, y_i, y_i^{*})\}$ that is \emph{i.i.d.} with the potential testing set, where each $z_i = \mathcal{P}(emb(p_i, \mathcal{R}_i))$ is a latent representation and $y_i^{*}$ is the formal verification result from $\mathcal{M_C}$.  
For each correctly classified sample ($y_i = y_i^{*}$), we compute its distance to the centroid $c^*$ of the corresponding cluster:
\begin{equation}
d_i = \| z_i - c^* \|_2, \quad c^* = \frac{1}{|\mathcal{S}|} \sum_{z_j \in \mathcal{S}} z_j, \nonumber
\end{equation}
where $\mathcal{S}$ is the set of correctly aligned (safe) samples.  
These distances form a calibration distribution $F_C(d)$ that captures how far safe samples typically lie from their centroid.

\textbf{Distance-Based Probabilistic Guarantee:}
For a new inference $(p', r'_{nl})$ with latent representation $z'$, we compute its distance to the nearest centroid. We then formulate a calibration distribution that captures the relation between the distances and prediction accuracies. This distribution is used to estimate the probability that a prediction $y'$ is correct, denoted as the \emph{probabilistic guarantee} $\hat{p}(y' \mid z')$.

\textbf{Definition 2 (Distance to Centroid):}
\label{def:centroid-repv}
Let $c_s$ and $c_u$ be the centroids of the safe and unsafe clusters in $\mathcal{Z}$.  
For a new input $(p', r'_{nl})$ with embedding $z'$, the distance to its nearest centroid is
\[
d' = \min\big( \| z' - c_s \|_2,\, \| z' - c_u \|_2 \big).
\]

\textbf{Definition 3 (Calibration Distribution):}
\label{def:calib-repv}
Let $F_C: \mathbb{R}^+ \rightarrow [0,1]$ denote the cumulative distribution of distances from correctly classified samples to their cluster centroids. Let $y_c^* \in \{0, 1\}$ be the safety label associated with $c^*$.
For a new input, $F_C(d')$ gives the probability that a sample from the opposite class lies beyond distance $d'$:
\[
F_C(d') = \Pr\big[ \| z_i - c^* \|_2 > d' \,\big|\, y_i \ne y_c^* \big].
\]

If the safe and unsafe classes are imbalanced, we compute $F_C$ separately for each class and select the corresponding $F_C$ based on the nearest centroid.

\textbf{Theorem 1 (Probabilistic Guarantee):}
\label{thm:repv-prob}
Given a new input with latent representation $z'$ and nearest centroid $c^* \in \{c_s, c_u\}$, the probabilistic guarantee $\hat{p}(y' \mid z')$ that its classification is correct is
\begin{equation}
\hat{p}(y' \mid z') = 1 - \frac{ (1 - F_C(d')) \cdot \Pr[y_i \ne y_c^*] }{ \Pr[\, \| z_i - c^* \|_2 \le d' \,] }.
\end{equation}

\textbf{Proof 1:}
Let $c^* \in \{c_s, c_u\}$ be the nearest centroid of the new input’s latent representation $z'$, and let $d' = \|z' - c^*\|_2$.  
By Definition 3, the calibration distribution satisfies
\[
F_C(d') = \Pr[\,\|z_i - c^*\|_2 > d' \mid y_i \ne y_c^*\,].
\]
We define the probabilistic guarantee as
\begin{align}
    \hat{p}(y' \mid z') & = \Pr[\,y_i = y_c^* \mid \|z_i - c^*\|_2 \le d'\,] \nonumber \\
    & = 1 - \Pr[\,y_i \ne y_c^* \mid \|z_i - c^*\|_2 \le d'\,]. \nonumber
\end{align}
Applying Bayes’ rule,
\begin{align}
\Pr[\,y_i \ne y_c^* & \mid \|z_i - c^*\|_2 \le d'\,] \nonumber \\
= & \frac{\Pr[\,\|z_i - c^*\|_2 \le d' \mid y_i \ne y_c^*\,] \Pr[\,y_i \ne y_c^*\,]}
{\Pr[\,\|z_i - c^*\|_2 \le d'\,]} \nonumber \\
= & \frac{(1 - F_C(d')) \Pr[\,y_i \ne y_c^*\,]}
{\Pr[\,\|z_i - c^*\|_2 \le d'\,]}. \nonumber
\end{align}
Substituting into the definition of $\hat{p}(y' \mid z')$, we get
\[
\hat{p}(y' \mid z') = 1 - \frac{(1 - F_C(d')) \Pr[\,y_i \ne y_c^*\,]}
{\Pr[\,\|z_i - c^*\|_2 \le d'\,]}.
\]
Hence proven.

\textbf{Verification:}
$\hat{p}(y' \mid z')$ denotes uncertainty associated with the rule compliance prediction. A higher $\hat{p}(y' \mid z')$ implies stronger alignment between linguistic interpretation and formal verification. 
Formally, a new plan $p'$ \textbf{complies} with a natural-language rule $r_{nl}'$ with a \textbf{guarantee} $\hat{p}(y' \mid z')$ if
\begin{align}
\label{eq: comply}
p' \models r_{nl}' & \iff \nonumber \\
\big[(\hat{y}'_{safe} & = 1 \wedge y' = 1) \ \lor\ (\hat{y}'_{safe} = 0 \wedge y' = 0)\big].
\end{align}

\section{Guarantee-Driven Refinement}
\label{sec: refine}

To close the loop between verification and planning, \texttt{RepV} uses the probabilistic guarantee $\hat{p}(y \mid z)$ as a feedback signal to refine the planner, which is a foundation model that produces plans. We adopt two complementary strategies: supervised fine-tuning and preference optimization.

\textbf{Supervised Fine-Tuning (SFT) via Guarantee Filtering:}  
We collect a dataset of plan–rule pairs $(p, r_{nl})$ along with their interpreter predictions and latent probabilities. We then filter this dataset to retain only high-confidence examples, i.e.\ those where the probabilistic guarantee exceeds a threshold $\tau$ (e.g.\ $0.9$). Formally, define
\[
\mathcal{D}_{\text{hf}} = \{\, (p,r_{nl}) : \hat{p}(y \mid z) \ge \tau \,\}.
\]
We use $\mathcal{D}_{\text{hf}}$ to fine-tune the planner via cross-entropy loss, training it to generate plans that are rule compliant. Because the filtered set excludes high-uncertainty examples, the fine-tuning process focuses on reinforcing desired patterns.

\textbf{Probabilistic Guarantee for Preference Ranking:}
We can use the probabilistic guarantee $\hat{p}(y \mid z)$ also as a measure of relative preference between alternative plans generated by the planner.

For each given task, the planner generates two different plans $p_1$ and $p_2$ by varying the random seed. Then, we estimate the guarantee for both plans: $\hat{p}(y_1 \mid z_1)$ and $\hat{p}(y_2 \mid z_2)$. We select the plan with \textbf{higher guarantee of rule compliance} as the \emph{preferred output}, and the other as the \emph{non-preferred output}. Lastly, we add this triplet (task prompt, preferred output, non-preferred output) into the fine-tuning dataset $\mathcal{D}_{\text{pr}}$.

We then directly pass $\mathcal{D}_{\text{pr}}$ into a \emph{Direct Preference Optimization} (DPO) framework to fine-tune the planner. Compared to supervised fine-tuning, preference optimization performs better when multiple correct answers exist.







%% file: tex/05-demonstration.tex
\section{Demonstration}
\label{sec: exp-setup}

We demonstrate \texttt{RepV} on navigation tasks using the Carla simulator \cite{dosovitskiy2017carla}.
In this section, we (1) present our experimental setting, (2) visualize the safety-separable latent space and its corresponding calibration distribution, and (3) present case studies that highlight its verification capability.

\begin{figure}[t]
    \centering
    \includegraphics[width=0.48\linewidth]{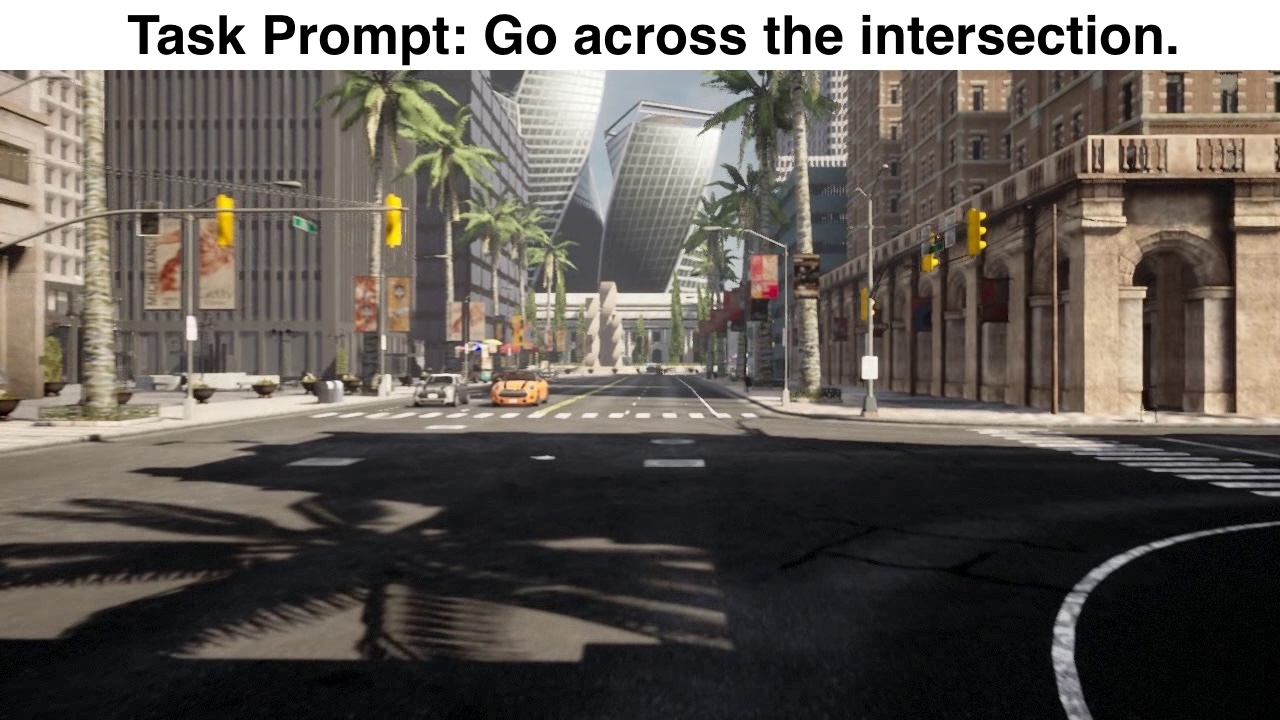}
    \includegraphics[width=0.48\linewidth]{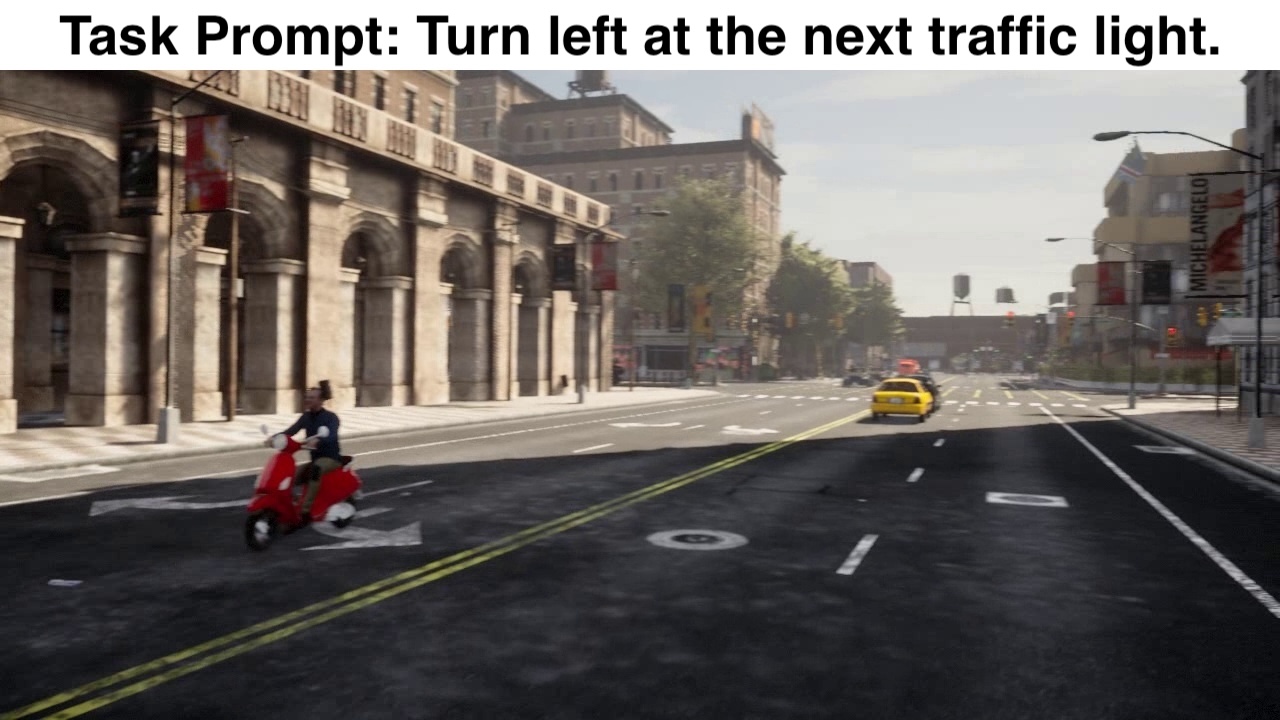}
    \vspace{-8pt}
    \caption{Example visual and textual input prompts to the planner.}
    \vspace{-10pt}
    \label{fig: prompt}
\end{figure}

\textbf{Experimental Setup: }
We design a series of navigation tasks, such as \textit{go straight}, \textit{turn left}, \textit{turn right}, \textit{park}, and \textit{make a U-turn at the traffic light or the stop sign intersection}. We use \texttt{GPT-o1-mini} as the interpreter and a \texttt{NuSMV} model checker~\cite{Cimatti2002NuSMV}. The projector comprises of three fully connected layers that map 1536-dimensional text embeddings (obtained from \texttt{text-embedding-3-small}) to a 10-dimensional latent space ($m = 10$).

Given a set of pre-implemented APIs, such as:
\begin{lstlisting}[style=nicePython]
# Publishes linear and angular velocity to the agent.
velocity_publisher(linear, angular)
stop() # Stops the robot.
# Returns a boolean value.
pedestrian_observed()
\end{lstlisting}
\vspace{-4pt}

\begin{figure*}[t]
\centering
    \includegraphics[width=0.24\linewidth]{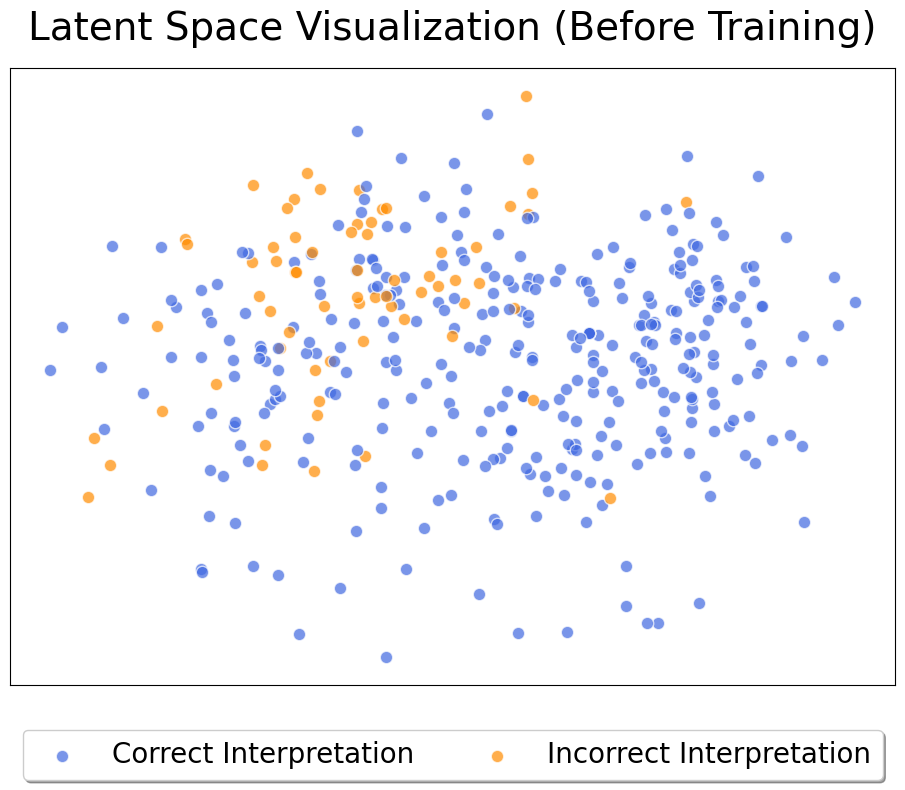}
    \includegraphics[width=0.24\linewidth]{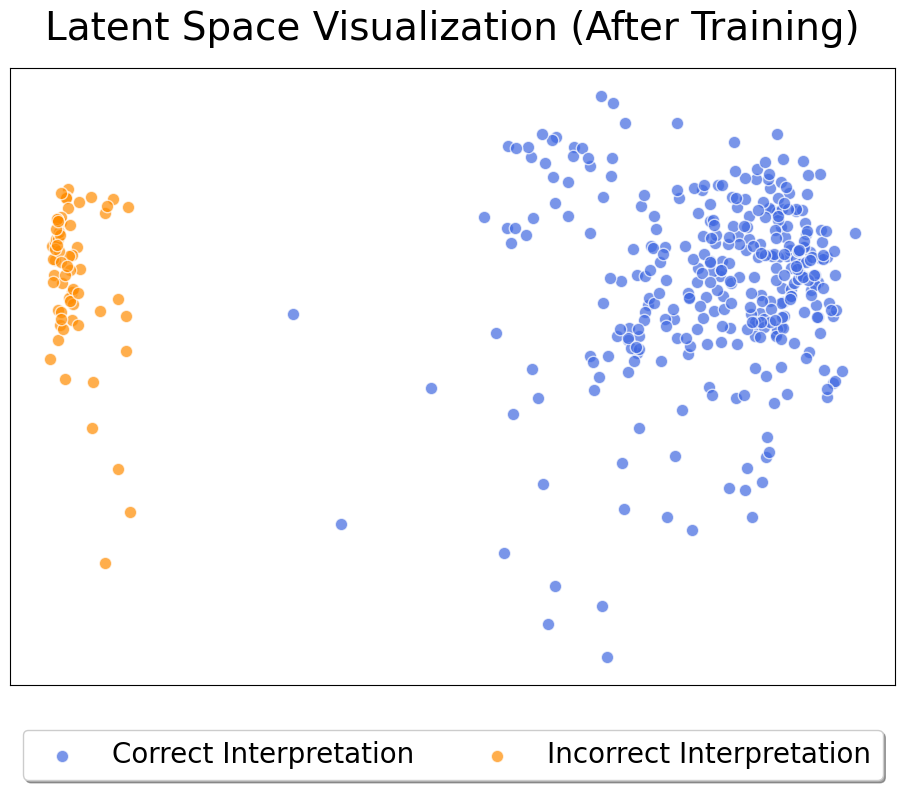}
    \includegraphics[width=0.24\linewidth]{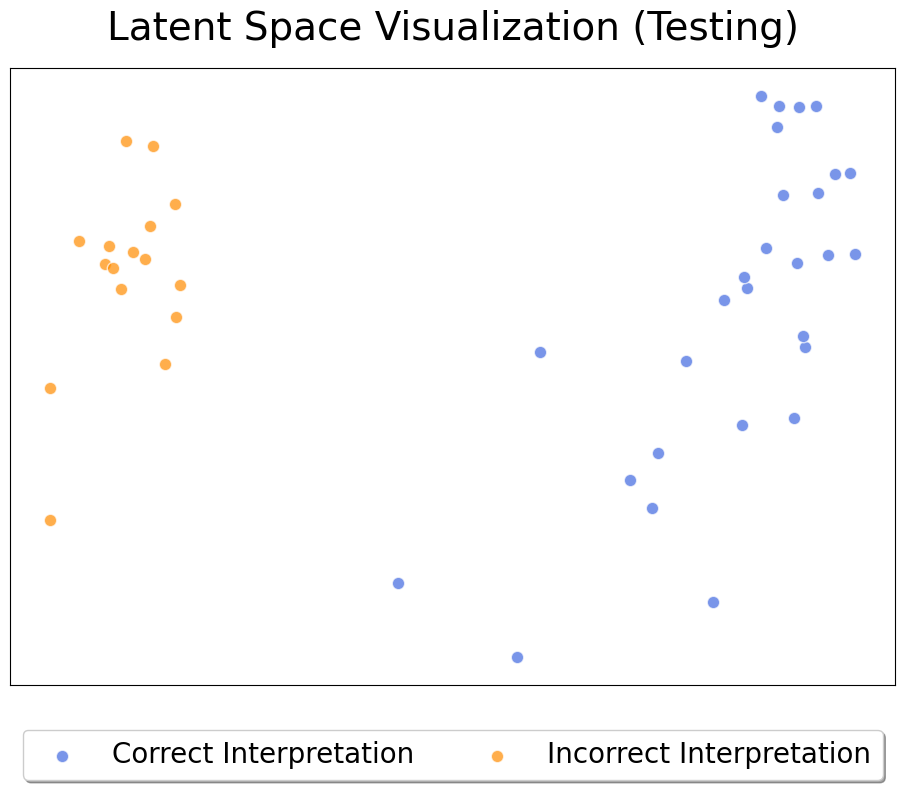}
    \includegraphics[width=0.24\linewidth]{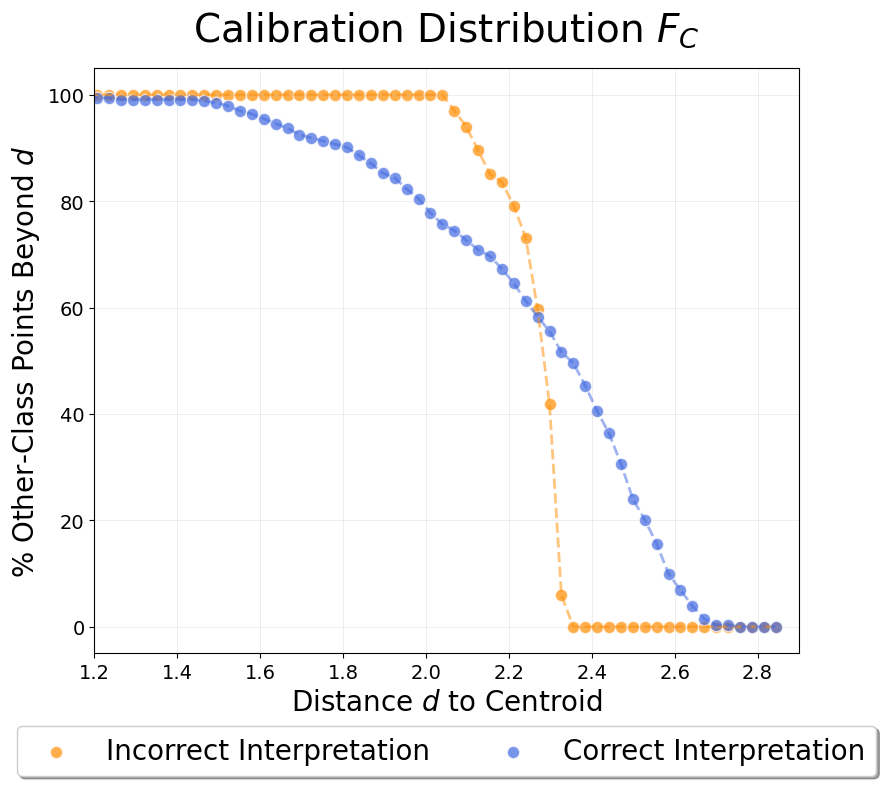}
\caption{The figures from left to right show: 1) the latent space before projector training, where safe and unsafe plans are inseparable, 2) latent space after training, which is separable by a linear decision boundary, 3) representations of \textit{out-of-domain} testing samples with \textit{new rules}, and 4) calibration distribution $F_C$ derived from a calibration set that drawn from the same distribution with the training samples.}
\vspace{-10pt}
\label{fig: latent}
\end{figure*}

we use \texttt{GPT-4.1-mini} as our planner to generate high-level plans composed of sequences of the system APIs. Specifically, the planner takes a text-based and/or image-based task prompt and the system APIs as inputs, producing a high-level plan, i.e., a Python program, that calls the APIs to accomplish the task. We show examples of input prompts in Figure \ref{fig: prompt} and output plans in Listing \ref{lst: code-test}.

\textbf{Training: } 
We generate 400 plans and 15 natural-language rules, each paired with its formal logic specification, to train the projector. For example, the rule \textit{“Yield to pedestrian”} corresponds to
\[
\phi_1 = \mathbf{G}\,\text{pedestrian} \rightarrow \mathbf{X}\neg\,\text{publish velocity}.
\]
The complete rules are in Listing~\ref{lst: rules} (Appendix~\ref{app: NL-rules}).

For each plan, we randomly sample one rule from the 15-rule set and query the interpreter to predict compliance and provide an explanation. Then, we use the model checker to obtain the ground-truth compliance label by verifying the corresponding formal specification. Each compliance prediction and rationale generation by the interpreter takes less than 3 seconds, while the model checking requires less than 0.1 seconds per plan.

After collecting 400 labeled samples, we train the projector, which contains \textbf{197 K parameters}, to predict rule compliance using cross-entropy loss with a batch size of 20 over 10 epochs. The full training process completes in \textbf{144 milliseconds}.
Figure~\ref{fig: latent} shows the latent spaces before and after training, demonstrating how the learned space achieves linear separability between safe and unsafe plans.

\textbf{Calibration: }
We construct a calibration set of 400 additional samples drawn from the same distribution as the training set, using the same set of natural-language rules and their corresponding formal specifications. 
Each sample is passed through the trained projector to obtain its latent representation and distance to the nearest cluster centroid. These distances form the calibration distribution $F_C$. 
This calibration process completes in only \textbf{225 milliseconds.} 
We show the resulting distribution in the right plot of Figure~\ref{fig: latent}.

\begin{figure}[t]
    \centering
    \input{figure/automata/automaton}
    \vspace{-10pt}
    \caption{Automaton-based representations of the plans for a Carla simulated driving task (left) and an indoor navigation task (right).}
    \label{fig: drive-automata}
\end{figure}
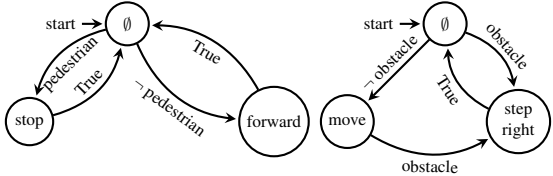

\textbf{Compliance Verification on Unseen Plans:}
Once the latent space and calibration distribution are established, we apply \texttt{RepV} to verify a new plan (Listing~\ref{lst: code-test}) against a new natural-language rule: \textit{Always avoiding pedestrians}.
This rule conveys the same underlying constraint as the training rules but is phrased differently, testing the model’s ability to generalize to paraphrased expressions of compliance.
\vspace{4pt}
\begin{lstlisting}[style=nicePython, label=lst: code-test, caption=A plan for the task: ``go across this intersection.'']
def go_across_the_intersection():
    ......
    while dist < 20.0:
      if pedestrian_observed():
        stop()
      else:
        velocity_publisher(speed, angle)
        dist = speed * (time.time() - t)
    ......
\end{lstlisting}
\vspace{-4pt}

Given the meta prompt to the interpreter (including the rule): \textit{``Does the code meet the rule: Always avoiding pedestrians?''} the interpreter classifies the plan as violating the rule ($y'=0$) with an explanation: 
\textit{``It partially meets the rule. It stops when a pedestrian is detected during turning and completes the turn if unobstructed.''}

Then, we obtain the latent representation of this plan $z'$. The distances of $z'$ to the centroids of the safe and unsafe clusters are $[1.2517,\, 1.0975]$, indicating that the sample lies closer to the unsafe cluster.  
The corresponding probabilistic guarantee is $\hat{p}(y' \mid z') = 0.971$, suggesting that \texttt{RepV} is confident that the interpreter's prediction $y'$ is incorrect. This inference procedure takes an average of \textbf{2.5 seconds}.

Hence, we conclude that the plan has a 97.1\% chance of satisfying the rule. To validate this conclusion, we convert the plan into an automaton (left in Figure~\ref{fig: drive-automata}) and verify it against $\phi_1$. The model checker confirms that the plan satisfies the specification.

This case highlights how \texttt{RepV} detects interpreter misclassifications. By leveraging the geometric separation and quantifying uncertainty in the latent space, \texttt{RepV} can rectify the language model interpreter's misclassification, demonstrating its ability to provide reliable, uncertainty-aware verification against natural-language rules.

%% file: figure/automata/automaton.tex
\begin{tikzpicture}[
    scale=.65,
    node distance=2.2cm,
    thick,
    every node/.append style={transform shape},
]

\node[state, initial] (q1)
    at (5.5,0)
    {\shortstack{$\emptyset$}};
\node[state] (q2)
    at (3.5,-2)
    {\shortstack{move}};
\node[state] (q3)
    at (7,-2)
    {\shortstack{step\\right}};

\path[->,sloped]

(q1) 
edge[] node[above]
    { $\neg$ obstacle }
    (q2)
edge[bend left] node[above]
    { obstacle }
    (q3)
    
(q2) 
edge[bend right] node[below]
    { obstacle }
    (q3)
    
(q3) 
edge[bend left] node[below]
    { True }
    (q1)
;

\node[state,initial] (q4)
    at (-1,0)
    {\shortstack{$\emptyset$}};
\node[state] (q5)
    at (-3,-2)
    {\shortstack{stop}};
\node[state] (q6)
    at (2,-2)
    {\shortstack{forward}};

\path[->,sloped]

(q4) 
edge[bend right] node[below]
    { pedestrian }
    (q5)

edge[bend right] node[below]
    { $\neg$ pedestrian }
    (q6)

(q5) 
edge[bend right] node[above]
    { True }
    (q4)

(q6) 
edge[bend right] node[below]
    { True }
    (q4)
;
\end{tikzpicture}

%% file: tex/05-quantitative.tex
\section{Quantitative Analysis}
\label{sec: quant}

We quantitatively evaluate \texttt{RepV} across multiple robotic platforms to assess its verification accuracy and impact on downstream planning performance, e.g., how the verification outcome guides the planner refinement.

\begin{figure}[t]
    \centering
    \includegraphics[width=\linewidth]{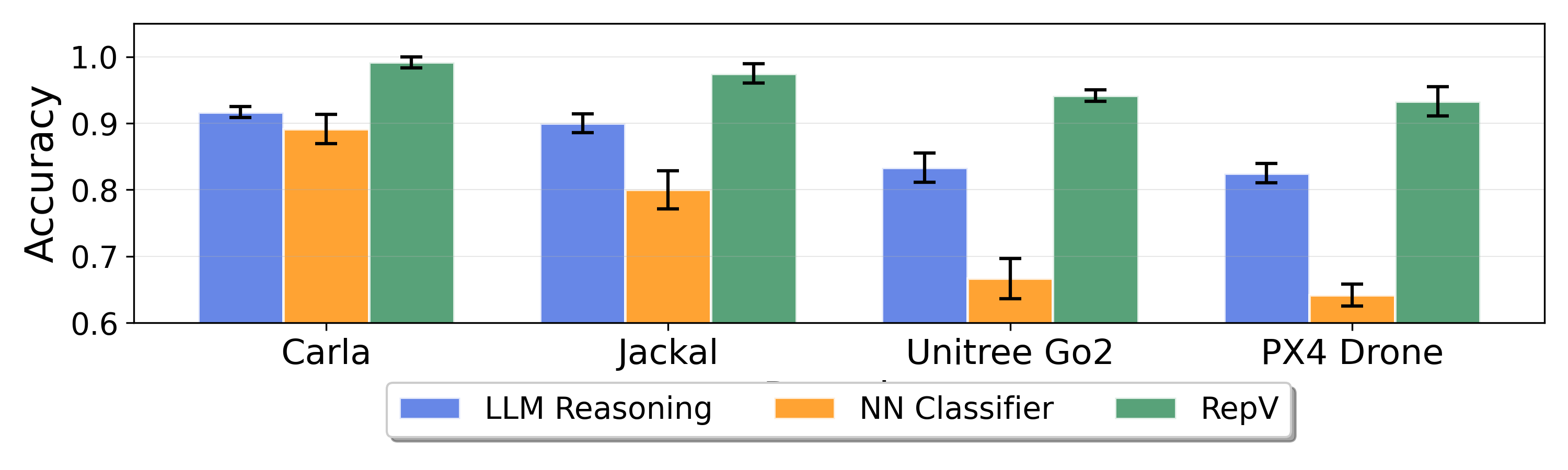}
    \vspace{-15pt}
    \caption{Comparison of three rule compliance prediction methods over four robot planning domains. RepV consistently achieves the highest accuracy compared to the two baselines. The error bar shows the standard deviation across five repetitive runs.}
    \label{fig: compliance-acc}
\end{figure}

\begin{table*}[t]
\centering
\caption{Fine-tuning Statistics. Acronyms: BS = Batch Size, LR = Learning Rate Multiplier, E = Epochs, TD = Training Duration (min), TT = Trained Tokens, DS = Data Size (KB), S = Number of Samples, S/E = Steps per Epoch, Steps = Total Steps, CE = Convergence Epoch, CStep = Convergence Step. We define convergence if the training compliance rate fluctuates by less than 1\% in the later epochs/steps. Our fine-tuning strategies \textbf{require fewer training samples and converge faster} than the baselines.}
\vspace{4pt}
\begin{tabular}{lccccccccccc}
\hline
\textbf{Fine-Tuned Model} & \textbf{BS} & \textbf{LR} & \textbf{E} & \textbf{TD} & \textbf{TT} & \textbf{DS} & \textbf{S} & \textbf{S/E} & \textbf{Steps} & \textbf{CE} & \textbf{CStep} \\
\hline
SFT (No Filter)   & 2 & 0.03 & 5 & 16.97 & 838{,}630 & 658.55 & 400 & 200 & 1000 & 5 & 1000 \\
SFT via Confidence Filter & 2 & 0.03 & 5 & 74.40 & 330{,}225 & 257.11 & 150 & 75 & 375 & 4 & 270 \\
SFT via Guarantee Filter (Ours) & 2 & 0.03 & 5 & 8.38 & 346{,}670 & 285.00 & 150 & 75 & 375 & 2 & 175 \\
DPO (Ours) & 2 & 0.03 & 5 & 50.68 & 370{,}760 & 233.22 & 200 & 100 & 500 & 2 & 120 \\
\hline
\end{tabular}
\label{tab:finetune-stats}
\end{table*}

\begin{figure*}[t]
    \centering
    \includegraphics[width=0.75\linewidth]{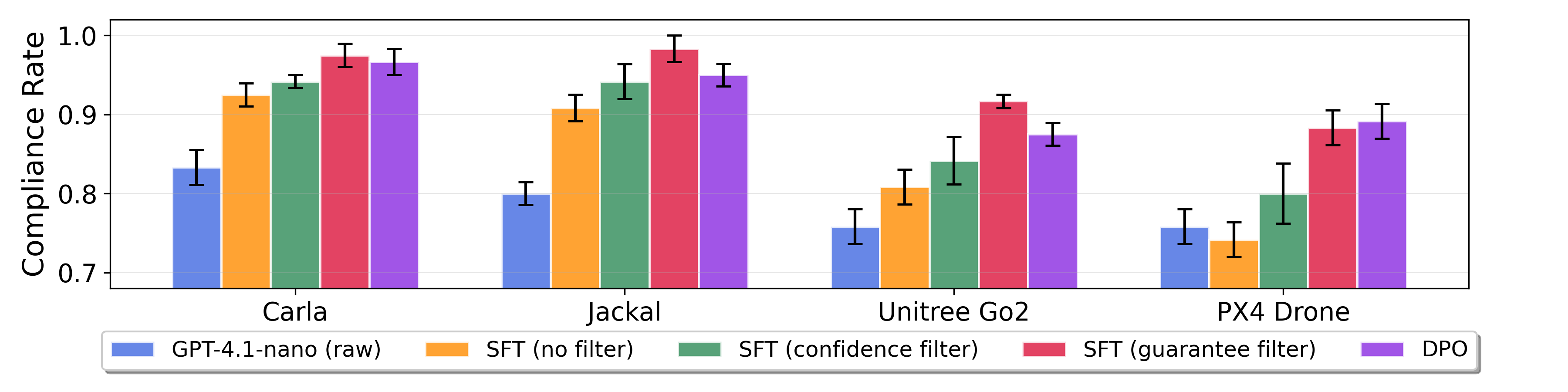}
    \hspace{-20pt}
    \includegraphics[width=0.25\linewidth]{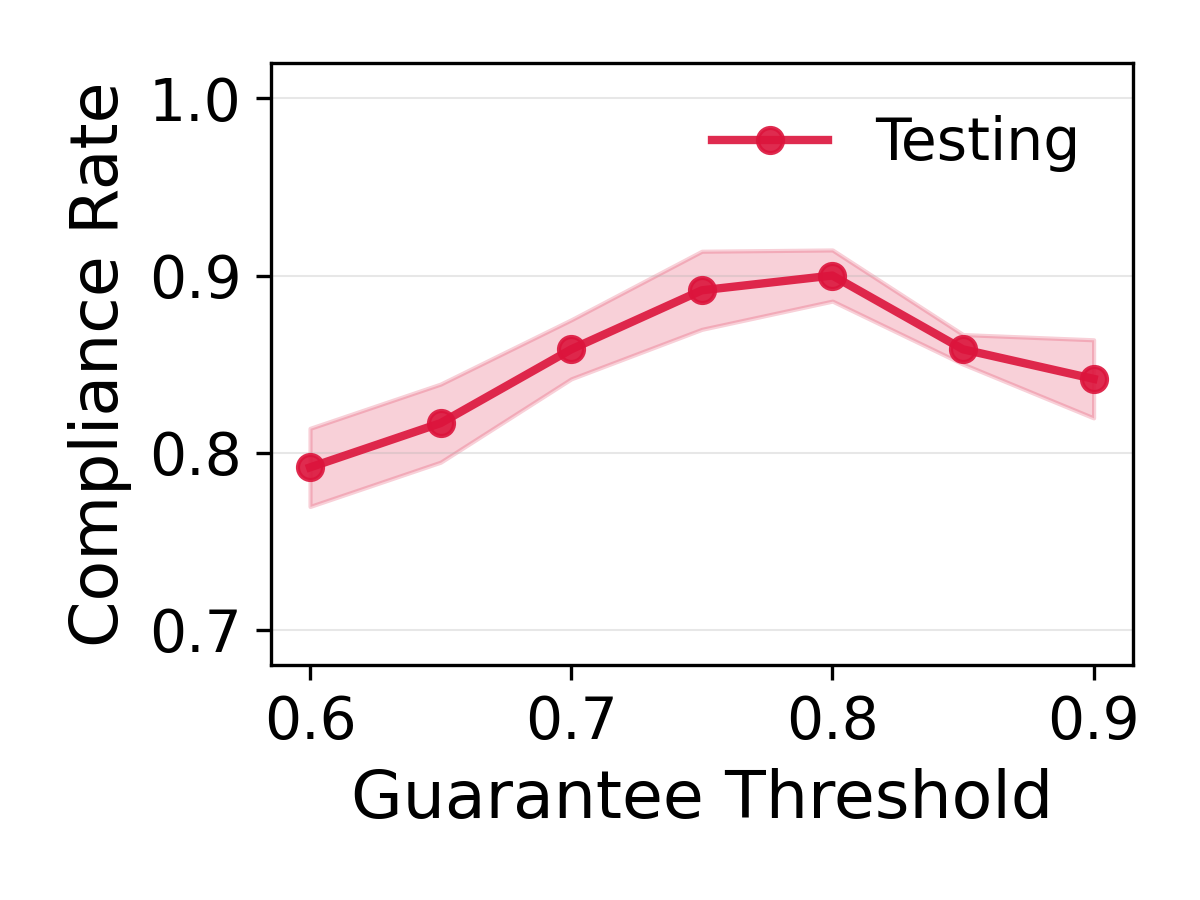}
    \vspace{-10pt}
    \caption{The left figure compares compliance rates across various refinement methods. The right figure shows the testing performance of our guarantee-filtered SFT at different guarantee thresholds, in which achieves the best the performance at threshold 0.8. The error bar shows the standard deviation across three repetitive runs. Our methods ({\color{red}\textbf{red}} and {\color{figurepurple}\textbf{purple}}) outperforms the other baselines in all robotic domains and achieves higher performance gain in out-of-domain tasks (Go2 and PX4 drone).}
    \vspace{-10pt}
    \label{fig: refine-acc}
\end{figure*}

\subsection{Rule Compliance Prediction}
\label{sec: compliance-inf}
We evaluate the compliance prediction accuracy of \texttt{RepV} across four application domains and compare it against two baseline methods, as shown in Figure~\ref{fig: compliance-acc}. 
Given a set of plans and a set of rules, we define \textit{compliance accuracy} as
\begin{equation}
\label{eq:acc}
\text{Accuracy} = \frac{1}{N}\sum_{i=1}^{N} \mathbb{I}\big[y_i = y_i^*\big],
\end{equation}
where $N$ is the number of testing samples, $y_i$ denotes the predicted compliance label, $y_i^*$ the ground-truth label from model checking, and $\mathbb{I}[\cdot]$ the indicator function.

\textbf{Baselines:} We compare our method against two baseline methods listed below.

(1) \textit{LLM Reasoning:} We follow the framework proposed by \cite{guan2024deliberative}, using a foundation model (\texttt{GPT-4.1-nano}) directly classifies plan compliance from text inputs via chain-of-thought reasoning.

(2) \textit{NN Classifier:} a 10-layer perception trained with 400 samples for 10 epochs, taking plan embeddings as input to predict compliance.  

(3) \textit{RepV (ours):} the proposed neurosymbolic verifier, which infers compliance according to Equation~\ref{eq: comply}.

\textbf{Domains:} We evaluate the baselines over four robot planning domains of increasing embodiment diversity:
\texttt{Carla} driving simulator, \texttt{Jackal} ground robot sharing identical APIs and tasks with Carla, \texttt{Unitree Go2} legged robot we used for indoor navigation, and a \texttt{PX4 Vision 1.0} quadcopter (drone) we used for 3D aerial navigation. Each domain contains 40 generated plans and 5 natural language rules, where each plan is checked against 5 rules separately. We present the rules in Appendix \ref{app: rules}.

\textbf{Results:}
\texttt{RepV} consistently achieves the highest compliance accuracy across all domains, maintaining at least 95\% compliance accuracy on in-domain tasks (Carla and Jackal) and 90\% accuracy on out-of-domain tasks (Go2 and Drone).  
In contrast, the two baselines exhibit significant degradation when transferred to new domains, highlighting the robustness of our verifier.

Figure \ref{fig: latent} (the third from left) visualizes the latent-space distributions of out-of-domain tasks (Go2 and Drone). Notably, these out-of-domain representations exhibit approximately identical distributions as those trained on Carla, showing that the latent space primarily captures the interpreter’s reasoning behavior rather than domain-specific information. Because the interpreter’s linguistic reasoning pattern remains consistent across various domains, the latent geometry of safe versus unsafe plans remains stable, enabling \textbf{robust cross-domain transfer without retraining}.

\begin{figure*}[t]
    \centering
    \includegraphics[width=0.49\linewidth]{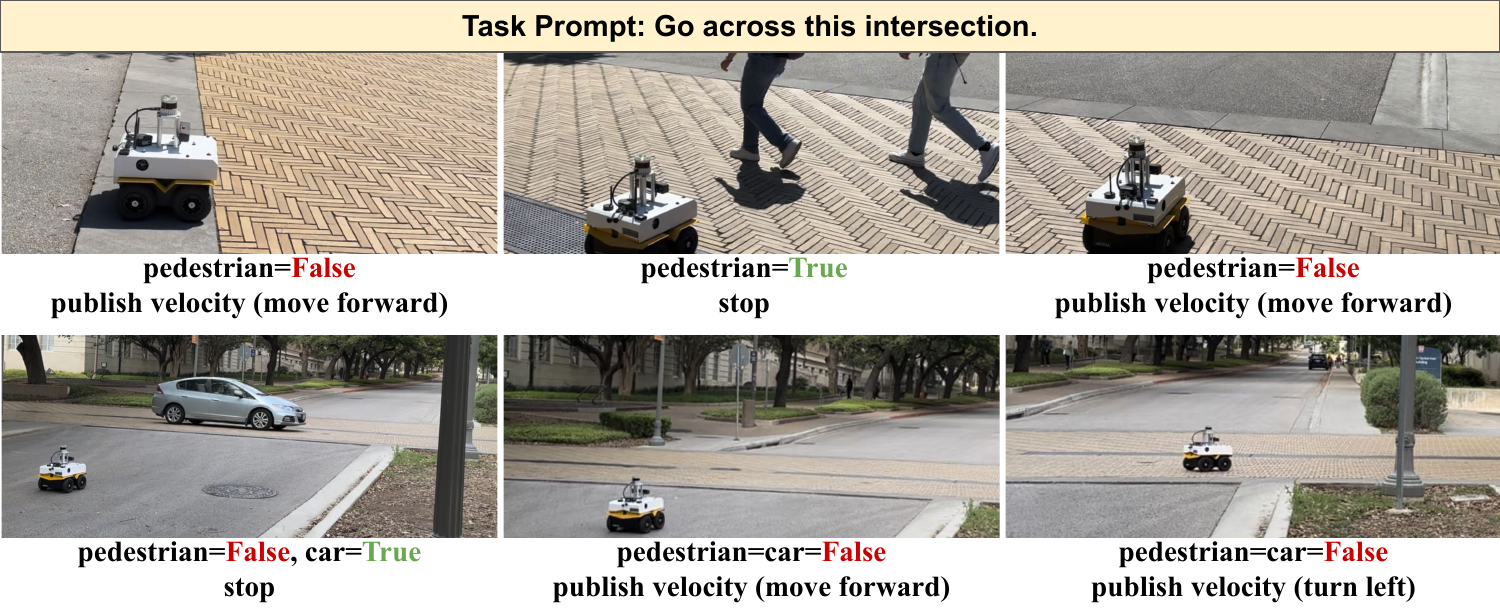}
    \includegraphics[width=0.49\linewidth]{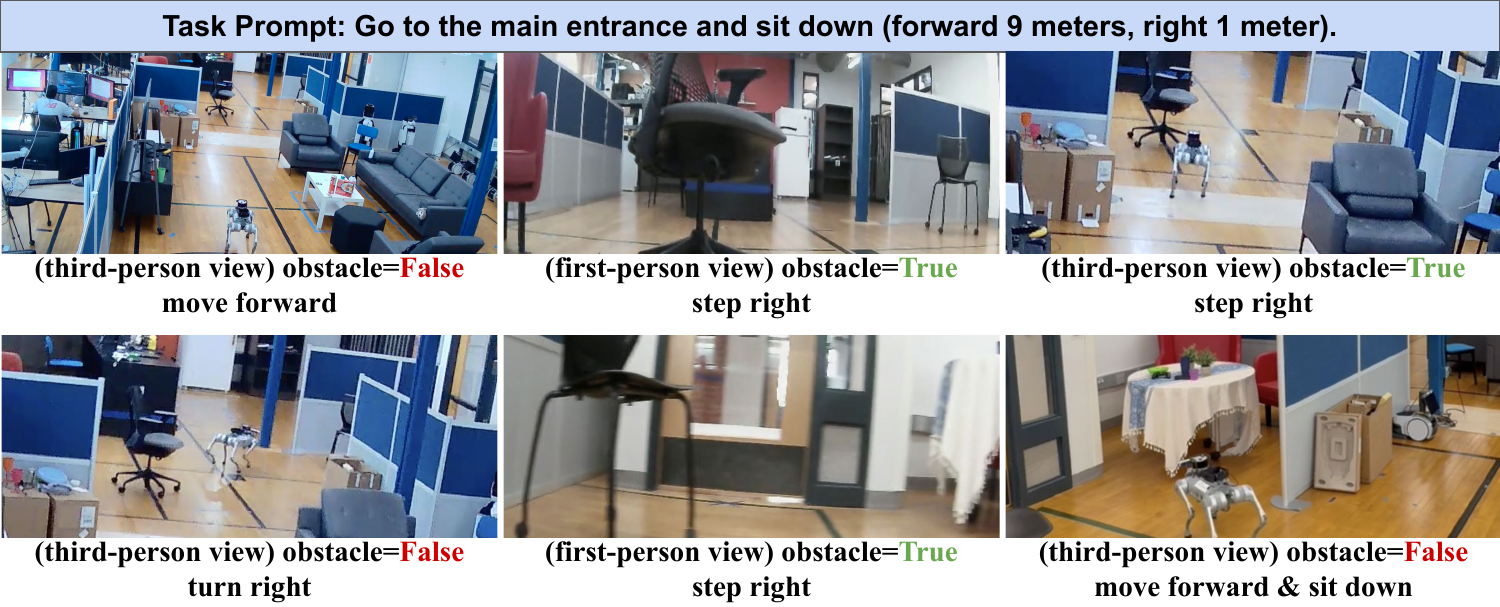}
    \vspace{-10pt}
    \caption{We execute the plan generated by our fine-tuned planner on real robots for outdoor and indoor navigation. The robots complete the task while complying with the domain-specific rules.}
    \vspace{-10pt}
    \label{fig: robot-demo}
\end{figure*}

\begin{figure}[t]
    \centering
    \includegraphics[width=\linewidth]{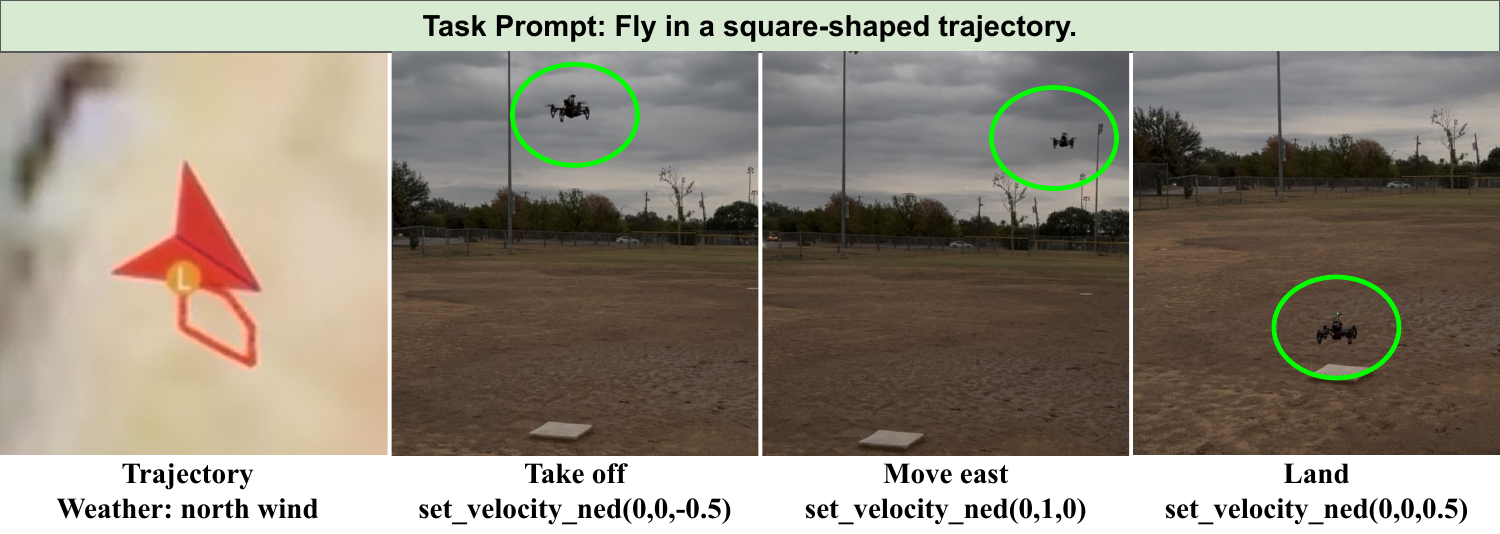}
    \vspace{-20pt}
    \caption{The fine-tuned planner generates executable plans that complies rules in aerial navigation.}
    \vspace{-10pt}
    \label{fig: drone}
\end{figure}

\subsection{Planner Refinement}

After demonstrating \texttt{RepV}’s capability in rule compliance prediction, we leverage its verification results to refine the knowledge-source planner. 
This section demonstrates how our guarantee-driven refinement effectively and efficiently enhances the planner to generate rule-compliant plans.

For quantitative analysis, we compare our guarantee-driven refinement methods against several baselines:

\textbf{SFT (No Filter):} Standard supervised fine-tuning using all foundation model-generated plan–rule pairs, regardless of their verification confidence.

\textbf{SFT via Confidence Filter:} Fine-tuning performed only on plan–rule pairs whose interpreter assigns a high softmax confidence score ($\ge 0.8$) to its compliance prediction.

\textbf{SFT via Guarantee Filter (Ours):} Fine-tuning restricted to samples satisfying the rules with probabilistic guarantee $\hat{p}(y \mid z) \ge \tau = 0.8$. We present the performance over different guarantee values in Figure \ref{fig: refine-acc} (right).

\textbf{DPO (Ours):} DPO using the probabilistic guarantee as a ranking signal between alternative plans.

To evaluate their performance, we define the metric \emph{compliance rate} as
\begin{equation}
\label{eq:compliance-rate}
\text{compliance rate} = \frac{1}{N} \sum_{i=1}^{N} \mathbb{I}\big[p_i \models r_{nl,i}\big],
\end{equation}
where $N$ is the number of testing plans, $p_i$ denotes the $i$-th generated plan, $r_{nl, i}$ the corresponding natural-language rule, and $\mathbb{I}[\cdot] = 1$ if and only if the plan is rule-compliant. This metric measures the proportion of generated plans that formally satisfy the given rules.

We present the fine-tuning details in Table~\ref{tab:finetune-stats} and evaluate the fine-tuned planners on 160 test samples, each labeled by the model checker using handcrafted formal specifications. The evaluation covers four domains: the Carla simulator (training environment, in-domain), the Jackal ground robot (real-world counterpart of Carla, sim2real transfer), the Unitree~Go2 (out-of-domain), and the PX4 (out-of-domain). We present the rules for each domain in Appendix \ref{app: rules}.

Figure \ref{fig: refine-acc} shows the compliance rates achieved by each planner. Across all domains, both of our fine-tuned models outperform the baselines by \textbf{10\% to 20\% in compliance rate}, and the performance gap widens in out-of-domain tasks. Meanwhile, our fine-tuning strategies \textbf{halve the number of training samples and the convergence time}.
These results highlight that using \texttt{RepV}'s probabilistic guarantee as feedback enables data-efficient, safety-aligned planner refinement and generalization across different tasks and rules.

\subsection{Real-World Deployment}
\label{sec: realworld}

We further validate the refined planners (\emph{guarantee-filtered}) in real-world environments to examine whether \texttt{RepV} can guide safe and executable behaviors. We present the complete task prompts and executable plans in Appendix \ref{app: plans}.

\textbf{Outdoor and Indoor Navigation:}
Figure \ref{fig: robot-demo} shows the execution of the generated and \texttt{RepV}-verified plans on two physical platforms. All the executed plans are verified as compliant with a probabilistic guarantee of 90\%+.

For the Jackal ground robot, we specify a rule \emph{``Yield to pedestrians and coming vehicles.''} The robot crosses the intersection and crosswalk while dynamically adjusting its actions to give right of way to pedestrians and nearby cars.

For the Unitree~Go2, we specify a rule that is distinct from outdoor navigation: \emph{``Bypass the obstacles; do not stop.''}
The resulting behavior, as presented in Figure \ref{fig: drive-automata} (right), demonstrates that the planner generalizes to different locomotion modalities and physical dynamics, achieving reliable obstacle avoidance without halting unnecessarily.

\textbf{Aerial Navigation:}
Figure \ref{fig: drone} extends this evaluation to a 3D aerial navigation task. The rule is: \emph{``Keep altitude below 10 meters and landing speed within 1~m/s.''}
The drone follows a high-level plan that draws a square trajectory under mild wind disturbance while respecting altitude and landing-speed constraints.

\textbf{Summary: } The RepV-refined planners successfully produce executable and rule-compliant plans in various physical environments, enabling \textbf{platform agnostic} and \textbf{safety-constrained plan generation and verification} across different robot embodiments and operational constraints.

%% file: tex/06-conclusion.tex
\section{Conclusion}

We introduce \texttt{RepV}, a neurosymbolic verifier that learns a \emph{safety-separable latent space} bridging the rigor of formal verification with the accessibility of natural-language reasoning. By embedding plans and interpreter rationales into this latent representation, \texttt{RepV} enables \emph{probabilistic-guaranteed compliance verification} without handcrafted logic specifications. 

Across simulated and real-world robotic applications, \texttt{RepV} achieves over 90\% compliance verification accuracy. Furthermore, we leverage the probabilistic verification outcome to guide planner refinement, improving compliance rates of the generated plans by 10--20\% while halving convergence time during refinement.

\textbf{Limitations and Future Work:} Although RepV generalizes across embodiments, the learned projector currently requires retraining when transferred to distinct reasoning domains beyond robotics and planning, such as medical or finance. On the other hand, we show that the projector learning and calibration require less than one second, making adaptation practical for new environments. Hence, the primary difficulty lies in projector training and calibration data collection, which may require domain expertise to provide formal or domain-specific constraints.

Moreover, \texttt{RepV} focuses on natural-language rules. Future works will aim to verify \emph{multimodal constraints} (visual traffic signals, speed-limit signs, or spatial safety boundaries) by integrating perception modules into the latent-space reasoning loop. Another direction is to apply \texttt{RepV} to other sequential decision-making domains like logistics, healthcare, and financial compliance, where natural-language rules work alongside formal specifications.

%% file: tex/A-appendix.tex
\onecolumn
\appendix

\section{Definitions}
\label{app: definition}

\textbf{Definition A1:}
\label{def: transition-system}
    A \textbf{\textsc{transition system}} $TS = (Q_{s}, T_{s}, L_{s})$ is a tuple of a set of states $Q_{s}$, a set of transitions $T_{s} = \{(q_i, q_j)\ |\ $ $ q_i, q_j \in Q_{s}\}$, i.e., $(q_i, q_j)$ means a transition from state $q_i$ to $q_j$, and a label function $L_{s}: Q_{s} \rightarrow 2^{AP}$.

$AP$ is a set of atomic propositions. Each atomic proposition has a truth value---true or false---but does not contain any logical connectives like "and," "or," "not," etc. 

\textbf{Definition A2:}
\label{def: automaton}
    A \textbf{finite-state automaton (FSA)} $\mathcal{A} = (Q_a, p_0, $ $ T_a, L_a)$ is a tuple consisting of a set of states $Q_a$, an initial state $p_0$, a set of transitions $T_a = \{(p_i, \sigma, p_j) \ |\ p_i, p_j \in Q_{a}, \sigma \in 2^{AP}\}$, and a label function $L_a: Q_a \rightarrow 2^{AP}$.

\textbf{Definition A3:}
\label{def: product}
    Given an FSA $\mathcal{A}$ and a transition system $TS$, a \textbf{\textsc{product automaton}} $\mathcal{A} \otimes TS$, is a tuple $(Q, Q_0, T, L)$, where
    \begin{itemize}
        \item $Q = \{(p, q) \ |\ p\in Q_a, q \in Q_s\}$, $Q_0 = \{p_0\} \times Q_s$, 
        \item $T = \{((p, q), (p', q')) \ |\ p \in Q_a, q \in Q_s, (p, L_s(q), p') \in T_a, (q, q') \in T_s \}$,
        \item and $L((p, q)) = L_a(p) \cup L_s(q), \text{ where } p \in Q_a, q \in Q_s$.
    \end{itemize}

\paragraph{Temporal Logic}
Temporal logic is a formal language that expresses system (represented in FSA) properties that evolve over time. 
It extends propositional logic by including temporal operators, such as $\mathbf{F}$ (``eventually'') and $\mathbf{G}$ (``always''), which allow for reasoning about the system's temporal behaviors.
An LTL formula consists of
\begin{itemize}
    \item A set of atomic propositions.
    \item A set of temporal operators describes the system's temporal behavior.
    \item A set of logical connectives, such as negation ($\neg$), conjunction ($\wedge$), and disjunction ($\vee$), that can be used to combine atomic propositions and temporal operators.
\end{itemize}

\section{\texttt{L2A}: Text to Automaton}
\label{app: text2aut}

\begin{figure}
    \centering
    \includegraphics[width=0.4\linewidth]{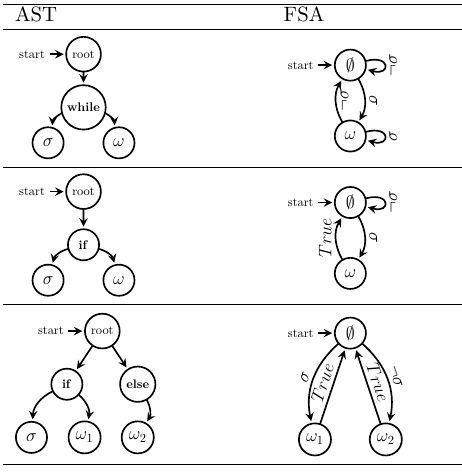}
    \caption{Keyword processor that converts particular grammars into automaton states and transitions.}
    \label{fig: keyword-handle}
\end{figure}

\begin{figure}
    \centering
    \includegraphics[width=0.8\linewidth]{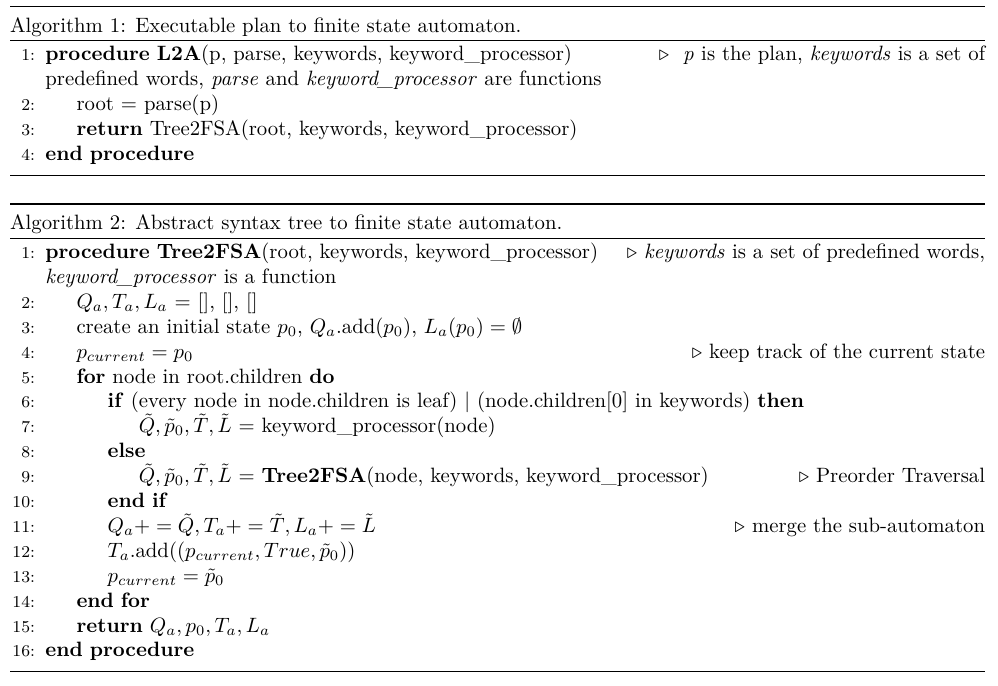}
    \caption{\texttt{L2A} algorithm that converts a text-based executable plan into a finite state automaton.}
    \label{fig: L2A}
\end{figure}

In our settings, we express the plan in programs, e.g., Python. The \texttt{L2A} algorithm first parses the plan into an abstract syntax tree (AST) using an existing library. Then, it converts the tree into an automaton as presented in Algorithm \ref{fig: L2A}. During this conversion, we define a set of keywords (e.g., Python keywords) and the conversion rules based on these keywords, as presented in Figure \ref{fig: keyword-handle}. When the algorithm observes a keyword with a predefined sentence structure, it follows the conversion rules defined in the keyword processor to translate the sentence into automaton states and transitions. Lastly, the algorithm composes all the states and transitions into a finite-state automaton.

\section{Empirical Study on Navigation Tasks}
\label{app: robot-nav}

We use the Carla simulator and the Jackal Clearpath robot for navigation tasks. They share the same set of APIs, but the former is executing in a simulated environment and the latter is executing in real world.

\subsection{Benchmark}
\label{app: benchmark}

As described in Section \ref{sec: compliance-inf}, the benchmark method \texttt{LLM-Reasoning} queries a language model for compliance classification.
We present the input prompts to the language model below.
\vspace{4pt}
\begin{lstlisting}[style=base]
Given the following plan in Python:

<Plan expressed in Python function>

Please analyze whether the code meets the rule: 
Do not remain stopped when the road is clear,

Please answer with a single letter 'Y' or 'N', indicating yes or no.
\end{lstlisting}

\subsection{Formal Logic Specifications}
\label{app: spec}
We present the complete list of specifications below.

$\phi_1 = \mathbf{G} \text{ pedestrian }\rightarrow \mathbf{X} \neg \text{ publish velocity }$,

$\phi_2 = \mathbf{F} \neg (\text{pedestrian} \vee \text{car}\vee \text{red light} ) \rightarrow \mathbf{X} \text{publish velocity}$,

$\phi_3 = \mathbf{G} (\text{ stop sign } \wedge \text{ car }) \rightarrow \mathbf{X} \neg \text{ publish velocity }$,

$\phi_4 = \mathbf{G}  \text{ stop sign} \rightarrow \mathbf{F} \text{ stop }$,

$\phi_5 = \mathbf{G}  \text{ red light} \rightarrow \mathbf{X} (\neg \text{ move forward } \wedge \neg \text{ turn left} )$.




We have APIs for observing stop signs, traffic lights, etc., that correspond to the atomic propositions ``red/green light,'' ``stop sign,'' etc. The propositions ``turn left/right,'' ``move forward,'' and ``stop'' correspond to the API \emph{velocity\_publisher} with different linear and angular speeds. For example, ``move forward'' corresponds to \emph{velocity\_publisher(linear=10, angular=0)}.

\subsection{Natural Language Rules}
\label{app: NL-rules}

As described in section \ref{sec: exp-setup}, we provide 30 natural language rules, from which we sample a subset of rules for training and evaluation. The rules are presented below.
\vspace{4pt}
\begin{lstlisting}[caption=Natural language rules for robot navigation, label=lst: rules]
============Training and Calibration======
Give the right of way to all pedestrians.
Let pedestrians pass first.
Always allow pedestrians to cross before proceeding.

Do not remain stopped if the road is clear.
Keep moving when it is safe to do so.
Avoid unnecessary stops when traffic allows you to proceed.

Stop where there is a vehicle ahead at a stop sign intersection.
Yield to oncoming cars at the stop sign.
Do not move ahead or turn if there is a vehicle at the stop sign intersection.

Make sure to stop at the stop sign.
Ensure a complete stop after seeing a stop sign ahead.
Always stop at the stop sign.

Always stop at a red light.
Do not proceed when the light is red.
Wait for the green light before moving.

======Testing======
Allow pedestrians to go first in all situations.
Pedestrians have the right of way, yield to them.
Slow down and let pedestrians cross safely.

Proceed when there are no obstacles.
Do not hesitate when the road is open.
Keep the flow of traffic moving when safe.

Give the right of way to approaching vehicles at the stop sign.
Let oncoming cars pass before proceeding from the stop sign.
Wait for approaching traffic to clear before moving past the stop sign.

Be sure to come to a complete stop at the stop sign.
Remember to halt when you reach the stop sign.
Do not forget to stop when you get to the stop sign.

Obey traffic signals, do not run red lights.
Do not cross the intersection on a red light.
Red means stop; never ignore it.
\end{lstlisting}

The first three rules from the training set and the first three from the testing set correspond to $\phi_1$ in \ref{app: spec}. The second group of 3 rules in both training and testing sets corresponds to $\phi_2$ in \ref{app: spec}. And the third, fourth, and fifth groups of 3 rules correspond to $\phi_3$, $\phi_4$, and $\phi_5$, respectively. During training and evaluation, we pass their corresponding formal logic specifications to the model checker to obtain the ground-truth compliance label.

\section{Empirical Study on Out-of-Domain Tasks}

\subsection{Rules and Specifications}
\label{app: rules}

\textbf{Indoor Navigation:}
We present the APIs for \texttt{Unitree Go2} legged robot below.
\begin{lstlisting}[style=nicePython]
obstacle_detected() # returns a boolean: detect an obstacle ahead of the robot
Move(self, vx: float, vy: float, vyaw: float) # vx is forward velocity and vy is velocity toward the right, vyaw is the counter-clockwise angular velocity
sleep(time_in_seconds: float)
StandUp()
StandDown()
\end{lstlisting}

For evaluation purposes, we use the model checker to verify the generated plan against the formal specifications presented below. The formal verification results serve as ground truth labels during evaluation.

$\phi_6 = \mathbf{G} \text{ obstacle }\rightarrow \mathbf{X} \neg \text{ forward }$,

$\phi_7 = \mathbf{G} \text{ obstacle }\rightarrow \mathbf{X} ( \text{ step left } \vee \text{ step right })$,

$\phi_8 = \mathbf{G} \neg\text{ obstacle }\rightarrow \mathbf{X} \neg \text{ stop }$,

the proposition ``obstacle'' corresponds to the API obstacle\_detected(); ``forward,'' ``step left'' and ``step right'' correspond to Move() with different input parameters; ``stop'' corresponds to Move(0,0,0,0).

During evaluation, we randomly select one of the following natural-language rules each time and pass it through our \texttt{RepV} framework for compliance verification. Then, we verify the corresponding formal logic specification to obtain the ground truth.
\vspace{4pt}
\begin{lstlisting}
============Corresponding to specification phi 6============
If an obstacle is detected, the agent must not move forward in the next time step.
The robot should never go forward immediately after sensing an obstacle.
Whenever there is an obstacle ahead, the next step is to avoid moving forward.

============Corresponding to specification phi 7============
If an obstacle is present, the next move should be a step to the left or right.
Upon detecting an obstacle, the agent must turn or sidestep, either left or right, on the next step.
Whenever an obstacle appears, the next action should be moving sideways.

============Corresponding to specification phi 8============
If there is no obstacle, the agent should keep moving.
Whenever the path is clear, the robot must continue rather than stop.
In the absence of obstacles, the next action must not be to stop.
\end{lstlisting}

\textbf{Aerial Navigation:}
We use the \texttt{PX4 Vision 1.0} quadracopter (drone) with the APIs presented below.
\begin{lstlisting}[style=nicePython]
set_velocity_ned(north, east, down, angle) # set drone velocity in m/s, the drone will keep moving in this velocity until it receives the next command
sleep_for(seconds: float) # the drone will keep the current action (velocity) during sleep time
obstacle_in_front() # returns a boolean
attitude_limit(max_meters)# returns a boolean, true if the drone does not exceed the limit
\end{lstlisting}

To obtain ground truth labels for quantitative analysis, we use the following two specifications:

$\phi_9 = \mathbf{G} (\text{ landing speed } < 1 )$,

$\phi_{10} = \mathbf{G} (\text{ attitude limit } )$,

where the proposition ``landing speed'' is captured by the third parameter of set\_velocity\_ned and ``attitude limit'' corresponds to the API attitude\_limit(max\_meters).

We present their corresponding natural language rules below.
\vspace{4pt}
\begin{lstlisting}
============Corresponding to specification phi 9============
The landing speed must stay below 1 m/s.
The drone should always maintain a landing speed below 1 meter per second.
The drone must never exceed a landing speed of 1 m/s.

============Corresponding to specification phi 10============
The maximum flying height is 5 meters.
The drone must always keep its height below 5 meters.
The drone should never rise higher than 5 meters at any time.
\end{lstlisting}

\subsection{Task Prompt and Executable Plans}
\label{app: plans}
Task Prompt for indoor navigation:
\vspace{4pt}
\begin{lstlisting}
Given the following drone APIs: 
obstacle_detected() $ returns a boolean: detects an obstacle in front of the robot
Move(self, vx: float, vy: float, vyaw: float) # vx is forward velocity and vy is velocity toward the right, vyaw is the counter-clockwise angular velocity
sleep(time_in_seconds: float)
StandUp()
StandDown()

Task: Go to the main entrance and sit down (forward 9 meters, right 1 meter).
Rule: Step right if there is an obstacle ahead.
\end{lstlisting}

We present the generated plan for the indoor navigation task below, which satisfies all the natural language rules with probabilistic guarantees above 90 percent. We present the real execution of the plan in Figure \ref{fig: robot-demo} (right).
\begin{lstlisting}[style=nicePython]
def go_to_main_entrance():
    # Step 1: Stand up from resting posture
    StandUp()

    # Parameters
    forward_speed = 0.5     # m/s forward
    sidestep_speed = 0.3    # m/s right
    turn_speed = -0.4       # rad/s yaw
    forward_distance_1 = 9  # meters
    forward_distance_2 = 1  # meters
    control_interval = 0.1  # seconds

    # Helper: move forward safely for given distance
    def move_forward(distance):
        duration = distance / forward_speed
        elapsed = 0.0
        while elapsed < duration:
            if obstacle_detected():
                # Try sidestepping right for 1 second to go around the obstacle
                for _ in range(int(1.0 / control_interval)):
                    Move(0.0, sidestep_speed, 0.0)
                    sleep(control_interval)
                # Resume forward motion
            else:
                Move(forward_speed, 0.0, 0.0)
            sleep(control_interval)
            elapsed += control_interval
        # Stop movement
        Move(0.0, 0.0, 0.0)

    # Step 2: Move forward 9 meters, avoid obstacles
    move_forward(forward_distance_1)

    # Step 3: Turn right
    turn_duration = math.abs(1.57 / turn_speed)
    elapsed = 0.0
    while elapsed < turn_duration:
        Move(0.0, 0.0, turn_speed)
        sleep(control_interval)
        elapsed += control_interval
    Move(0.0, 0.0, 0.0)

    # Step 4: Move forward 1 meters
    move_forward(forward_distance_2)

    # Step 5: Stop and stand down
    Move(0.0, 0.0, 0.0)
    sleep(0.5)
    StandDown()
\end{lstlisting}

Task Prompt for aerial navigation:
\vspace{4pt}
\begin{lstlisting}
Given the following drone APIs: 
set_velocity_ned(north, east, down, angle) # set drone velocity in m/s, the drone will keep moving in this velocity until it receives the next command 
sleep_for(seconds: float) # the drone will keep the current action (velocity) during sleep time 
obstacle_in_front() # returns a boolean 
attitude_limit(max_meters)# returns a boolean 

Task: Generate a plan for flying in a 4-meter square-shaped trajectory.
Rule: Do not exceed 5 meters in height. The landing speed must stay below 1 m/s.
\end{lstlisting}

We show the generated plan below. The plan satisfies all the natural language rules with guarantees above 95 percent. We safely execute the plan in physical environment and present the execution in Figure \ref{fig: drone}.
\begin{lstlisting}[style=nicePython]
def fly_square_4m():
    side_length = 4.0        # meters
    horizontal_speed = 1.0   # m/s
    vertical_speed = 0.5     # m/s
    max_altitude = 3.0       # meters (safety bound)
    control_interval = 0.1   # seconds for safety checks
    segment_time = side_length / horizontal_speed
    segment_time_vertical = max_altitude / vertical_speed

    def move_with_safety(north, east, down, angle, duration):
        elapsed = 0.0
        set_velocity_ned(north, east, down, angle)
        while elapsed < duration:
            if obstacle_in_front():
                # Stop immediately if an obstacle is detected
                set_velocity_ned(0.0, 0.0, 0.0, angle)
                return False
            if attitude_limit(max_altitude):
                # Stop if attitude or altitude exceeds safe bounds
                set_velocity_ned(0.0, 0.0, 0.0, angle)
                return False
            sleep_for(control_interval)
            elapsed += control_interval
        set_velocity_ned(0.0, 0.0, 0.0, angle)
        return True

    # Up
    if not move_with_safety(0.0, 0.0, -vertical_speed, 0, segment_time_vertical):
        return
    # East
    if not move_with_safety(0.0, horizontal_speed, 0.0, 0, segment_time):
        return
    # South
    if not move_with_safety(-horizontal_speed, 0.0, 0.0, 0, segment_time):
        return
    # West
    if not move_with_safety(0.0, -horizontal_speed, 0.0, 0, segment_time):
        return
    # North
    if not move_with_safety(horizontal_speed, 0.0, 0.0, 0, segment_time):
        return
    # Down and Stop
    set_velocity_ned(0.0, 0.0, vertical_speed, 0, segment_time_vertical)
    print("Completed square flight.")
    set_velocity_ned(0.0, 0.0, 0.0, 0, 1.0)
\end{lstlisting}